\documentclass[10pt,journal,compsoc]{IEEEtran}

\newcommand{\ncolor}{black}
\newcommand{\ncolorR}{black}

\usepackage{url}
\usepackage{balance}
\usepackage{epsfig}
\usepackage{graphicx}
\usepackage{amsmath}
\usepackage{amssymb}
\usepackage{xcolor}
\usepackage{algorithm}
\usepackage{algorithmic}
\newcommand{\isEq}[1]{\overset{#1}{\sim}}
\usepackage{multirow}
\usepackage[outercaption]{sidecap}
\usepackage{balance}
\usepackage[pagebackref=true,breaklinks=true,letterpaper=true,colorlinks,bookmarks=false]{hyperref}

\ifCLASSOPTIONcompsoc
  \usepackage[nocompress]{cite}
\else
  \usepackage{cite}
\fi
\ifCLASSINFOpdf
\else
\fi
\begin{document}
%
\title{Attack to Fool and Explain Deep Networks}
%
%
%
%

\author{Naveed~Akhtar*,~
        Mohammad A. A. K. Jalwana*, Mohammed Bennamoun, and
        Ajmal Mian
\IEEEcompsocitemizethanks{
\IEEEcompsocthanksitem *Authors claim joint first authorship based on equal contribution.
\IEEEcompsocthanksitem All authors are with the Department of Computer Science and Software Engineering, University of Western Australia, 35 Stirling Highway, 6009 Crawley, WA.\protect\\
E-mail: \{naveed.akhtar@ and mohammad.jalwana@research.\}uwa.edu.au,  \{mohammed.bennamoun and ajmal.mian\}@uwa.edu.au}
\thanks{Manuscript received June 2020, revised November 2020 and March 2021, accepted May 2021.}}

%
%

\markboth{Journal of \LaTeX\ Class Files,~Vol.~XX, No.~X, August~2020}%
{Shell \MakeLowercase{\textit{et al.}}: Bare Demo of IEEEtran.cls for Computer Society Journals}
%



\IEEEtitleabstractindextext{%
\begin{abstract}
Deep visual models are susceptible to adversarial perturbations to inputs. Although these signals are carefully crafted, they still appear noise-like patterns to humans. This observation has led to the argument that deep visual representation is misaligned with human perception. We counter-argue by providing evidence of human-meaningful patterns in adversarial perturbations. We first propose an attack that fools a network to confuse a whole category of objects (source class) with a target label. Our attack also limits the unintended fooling by samples from non-sources classes, thereby circumscribing human-defined semantic notions for network fooling. 
We show that the proposed attack
%
not only leads to the emergence of regular geometric patterns in the perturbations, but also reveals insightful information about the decision boundaries of deep models. Exploring this phenomenon further, we alter the `adversarial' objective of our {\color{\ncolor} attack} to use it as a tool to `explain' deep visual representation. 
We show that by careful channeling and projection of the perturbations computed by our method, we can visualize a model's understanding of  human-defined semantic notions. Finally, we exploit the explanability properties of our perturbations to perform image generation, inpainting and interactive image manipulation by attacking adversarialy robust `classifiers'. {\color{\ncolor} In all, our major contribution is a novel pragmatic adversarial attack that is subsequently transformed into 
%
%
a tool 
to interpret the visual models. The article also makes secondary contributions in terms of establishing the utility of our attack beyond the adversarial objective with multiple interesting  applications.}



\end{abstract}

\begin{IEEEkeywords}
Adversarial examples, perturbations, targeted attack, model interpretation, explainable AI.
\end{IEEEkeywords}}

\maketitle

\IEEEdisplaynontitleabstractindextext

%
\IEEEpeerreviewmaketitle

\IEEEraisesectionheading{\section{Introduction}\label{sec:introduction}}
\IEEEPARstart{A}{dversarial} examples~\cite{szegedy2013intriguing} are carefully manipulated inputs that appear natural to humans but cause deep models to misbehave.     
Recent years have seen multiple methods to generate the underlying manipulative signals (i.e.,~perturbations) for fooling deep models on individual input samples~\cite{szegedy2016rethinking}, \cite{goodfellow2014explaining}, \cite{moosavi2016deepfool}, \cite{dong2018boosting}, \cite{inkawhich2019feature} or 
many diverse samples
with a high probability~\cite{moosavi2017universal}, \cite{li2019universal} - termed `universal'  perturbations. 
The former may also  launch `targeted' attacks, where the model predicts the label of attacker's choice for  the input adversarial example.
The existence of adversarial examples is being  generally perceived as a threat to deep learning~\cite{akhtar2018threat}. Nevertheless,  controlled manipulation of model prediction  with input  perturbations also provides an opportunity of leveraging such signals for analyzing deep models. 

\begin{figure*}[t]
    \centering
    \includegraphics[width = 0.8\textwidth]{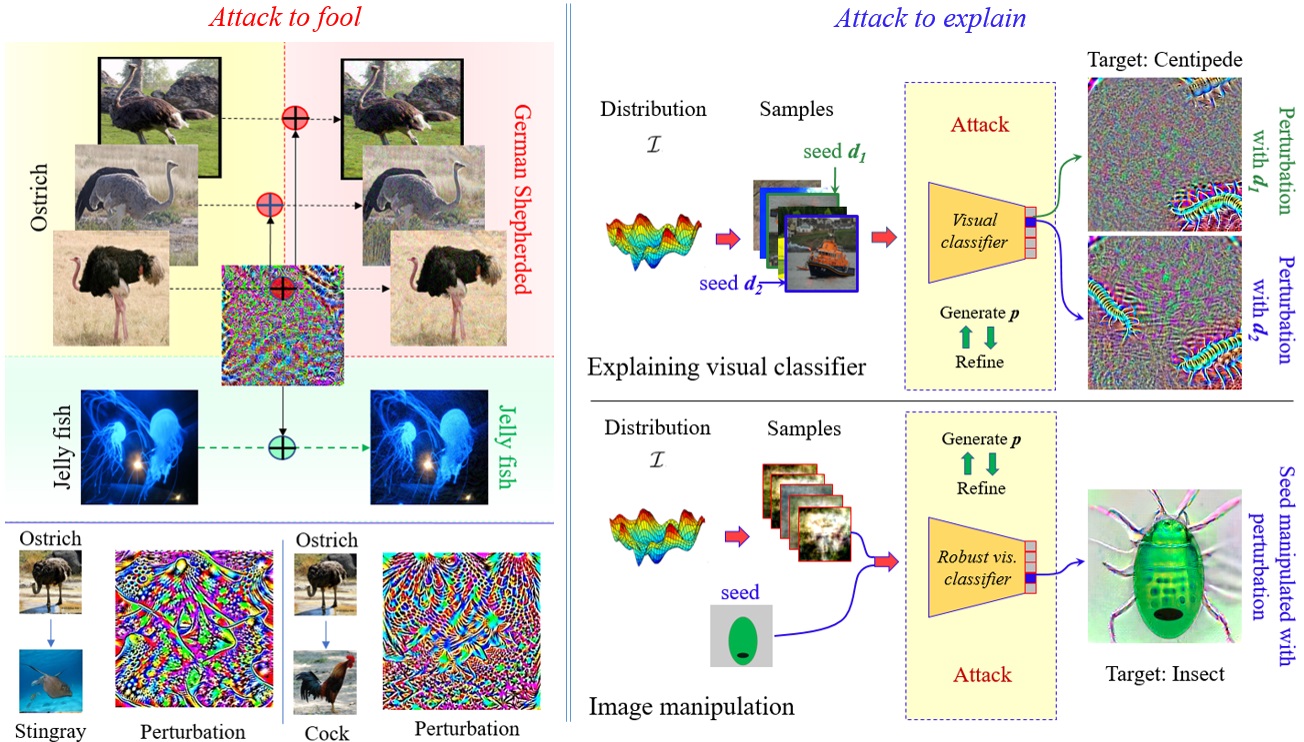}
    \caption{\textbf{Left}: (Top) A single perturbation alters the category of source class samples (Ostrich) to the target label of choice (German Shepherd) while inhibiting the influence of fooling on non-source classes (Jelly fish). The  adversarial examples are shown for ResNet-50~\cite{he2016deep}. (Bottom) A variant of the  perturbation reveals that the constituting signals  exploit correlation between the salient visual features of the incorrect class to fool deep models. The shown perturbations fool VGG-16~\cite{simonyan2014very}. \textbf{Right}: (Top) 
    Using an image distribution $\mathcal I$, we iteratively generate and refine  a perturbation $\boldsymbol{p}$ for a classifier to extract geometric patterns deemed salient for an object category  by the classifier. Anchoring the perturbation with different seed images $\boldsymbol{d}_n$ we can allow output variety. (Bottom)
    Attacking an adversarially `robust' classifier with our perturbations enables visually appealing image manipulation.}
    \label{fig:teaser}
    \vspace{-3mm}
\end{figure*}

In this work, we first introduce a technique to generate manipulative signals to fool deep models into confusing `an entire category of objects' with another target label of our choice, see Fig.~\ref{fig:teaser}(left:top). From the adversarial perspective, the resulting attack
is of high relevance in practical settings. It allows pre-computed perturbations that can change an object's category or a person's identity for a deployed model on-the-fly, where the attacker has also the freedom of choosing  the target label, and there is no particular constraint over the input. Concurrently, the convenient control over the manipulative signal in our attack encourages a fresh perspective of seeing adversarial perturbations as a deep model analysis tool.   
By appropriately leveraging the perturbation domain and restricting its range, our {\color{\ncolor} attack} reveals insightful patterns in the manipulative signal, as shown in Fig.~\ref{fig:teaser} (left:bottom).

Recently, Ilyas et al.~\cite{ilyas2019adversarial} claimed that existing large datasets (e.g.~ImageNet~\cite{deng2009imagenet}) admit to brittle yet highly predictive features that remain imperceptible to humans. 
It is argued that deep visual models rely on these non-robust features for high accuracy, which  also makes them susceptible to adversarial perturbations. 
Reliance of deep models on these apparently incomprehensible features is taken as an indication of misalignment between deep visual representation and human perception~\cite{engstrom2019learning}.
To remove this misalignment, Engstorm et al.~\cite{engstrom2019learning} proposed to learn deep models under a robust optimization framework. However, this entails a significant performance loss for the original model and a drastic increase in the computational complexity of model induction.  

We find it paradoxal that a representation misaligned with human perception still performs human-meaningful visual tasks with high accuracy. Thus, we leverage our systematically computed perturbations to delve deeper into the composition of these signals with an alternate objective of model `explanation' instead of model `fooling'.
We discover that it is possible to isolate human-understandable visual features of the target label by attacking the classifiers with {\color{\ncolor} this modification}, see Fig.~\ref{fig:teaser} (right:top). 
Within the context of adversarial perturbations, this observation weakens the argument of misalignment between human perception and deep representation. Rather, it places adversarial perturbations as human-meaningful geometric features of the target label, albeit in a primitive and subtle form.

In both cases of fooling and explaining a deep model, our perturbation estimation {\color{\ncolor}algorithms}  stochastically maximize the prediction probability of an image distribution's perturbed samples for a given target label. The maximization takes place  by iteratively stepping in the Expected gradient direction of the classifier's loss surface \textit{w.r.t.}~the input samples.
The optimization is guided by gradient moments to achieve the ultimate objective more efficiently.
For `fooling' a classifier, we restrict the input image distribution to the natural images of a single object category, termed the source class. Our technique additionally stops the leakage of fooling effects to non-source classes by carefully constraining the  perturbation.
For  model `explanation', the constraint over the input image distribution is loosened, and the search for perturbation is anchored by a seed image.   
We further channel the perturbation signal to focus more on its regions that cause  high activity of the neurons in the deeper layers of the classifier.
This procedure is purely based on the intermediate perturbations computed by our algorithm, which keeps our technique model faithful -  an appealing property for model explanation methods~\cite{engstrom2019learning}. The computed perturbation eventually provides visualization of the features attributed to the target label by the model.

Besides explaining deep models and highlighting the alignment of deep representation with human perception, the proposed attack also suits the paradigm of performing low-level vision tasks, e.g.~image generation, inpainting and interactive image manipulation, using robust `classifiers'~\cite{santurkar2019computer}, see Fig.~\ref{fig:teaser}(right:bottom).
We perform these tasks to affirm the utility of our technique (and  perturbations in general) beyond the adversarial objective by achieving significant visual improvements for these tasks over the original proposal in \cite{santurkar2019computer}. 
{\color{\ncolor} These are the secondary contributions that this article makes, besides the major contribution of a novel  adversarial attack  (\S~\ref{sec:AdvP}) that is systematically converted into an attack to explain deep visual models (\S~\ref{sec:ExpPComp}). The overall contributions can be summarized as}:
\vspace{-1mm}
\begin{itemize}
\item {\color{\ncolor}We propose a pragmatic adversarial attack} that is able to perform class-specific targeted fooling of deep visual classifiers, {\color{\ncolor} and modify it into an attack to interpret deep representation}. 
\item For model fooling, our targeted attack spans a whole object category instead of a single image and restricts the fooling effects to the source class.
\item We show effective targeted fooling of VGG-16~\cite{simonyan2014very}, ResNet-50~\cite{he2016deep}, Inception-V3~\cite{szegedy2016rethinking} and  MobileNet-V2~\cite{sandler2018mobilenetv2} on ImageNet~\cite{deng2009imagenet}, and ResNet-50 on the large-scale VGG-Face2 dataset~\cite{cao2018vggface2}. We also show that the computed adversarial examples transfer well to the Physical World.
\item {\color{\ncolor}With the attack to} explain a model, our perturbations manifest  visual features attributed by the model to individual object categories. The computed human-meaningful perturbations for `non-robust' classifiers weaken the argument  that deep representation is misaligned  with the human perception. 
\item We demonstrate visually appealing image generation, inpainting and interactive image manipulation by attacking robust classifiers. Our results affirm the utility of perturbations beyond model fooling.  
\end{itemize}

\vspace{-1mm}
This article is a considerable extension of our preliminary work presented in~\cite{ourCVPR}, which proposed an attack for model explanation. Here, we employ a broader framework that additionally allows pragmatic targeted fooling of a given model. It clearly  contextualizes  
our contribution within the adversarial attacks literature.
The newly introduced `adversarial' perspective of our technique is thoroughly explored on large-scale data  models, including experiments for the Physical World.
{\color{\ncolor} A parallel treatment of adversarial and explanation optimisation objectives with closely related algorithms}  offers clear insights into the potential of perturbations beyond the common adversarial utilities of these signals.  This article also extends the discussions in the preliminary work and reviews additional related literature. 

\vspace{-3mm}
\section{Related work}
Adversarial perturbations have attracted a considerable interest of computer vision community in the recent years~\cite{akhtar2018threat}.
These signals are commonly seen as a tool to fool deep learning models. Consequently, the existing  literature revolves around attacking deep models using input perturbations, or defending them against the resulting attacks. We first discuss the key contributions under this `adversarial' perspective of input perturbations, which divides the literature into the broad categories of adversarial attack and defense schemes. Later, we highlight the contributions that can be categorised as `non-adversarial' approaches, in the sense that the core objective of those methods deviates from the typical fooling/defending of deep  models.

\vspace{-3mm}
\subsection{Adversarial perspective}
\subsubsection{Adversarial attacks} 
Imperceptible additive input perturbations that can arbitrarily alter the decisions of deep visual models made their first appearance in the seminal work of Szegedy et al.~\cite{szegedy2013intriguing}.
This discovery led to numerous techniques of  fooling deep models. Goodfellow et al.~\cite{goodfellow2014explaining} devised the Fast Gradient Sign Method (FGSM) to efficiently compute  adversarial perturbations. The FGSM performs a single step gradient ascend over the loss surface of a model \textit{w.r.t.} the input. Later,  Kurakin et al.~\cite{kurakin2016adversarial} built on this concept by proposing a mutli-step Iterative FGSM (I-FGSM). The success of gradient based adversarial perturbations resulted in multiple follow-up algorithms, including Momentum I-FGSM  (MI-FGSM)~\cite{dong2018boosting},  Variance-Reduced I-FGSM (vr-IGSM)~\cite{wu2018understanding} and Diverse Input I-FGSM (DI\textsuperscript{2}-FGSM) \cite{xie2019improving} etc. 
Along the same line, Madry et al.~\cite{madry2017towards} noted that the `projected gradient descent on the negative loss function' strategy originally adopted by Kurakin et al.~\cite{kurakin2016adversarial} results in highly effective attacks. 
DeepFool~\cite{moosavi2016deepfool} is another popular attack that computes adversarial perturbations iteratively by linearizing the attacked model's decision boundaries near the input images.

The above-mentioned key contributions triggered a wide interest of the research community in adversarial attacks. More recently, this research direction is  seeing techniques for computing adversarial examples that transfer better to black-box models, and  computing perturbations with reduced norms for  imperceptibility~\cite{shi2019curls}, \cite{rony2019decoupling},
\cite{croce2019sparse},
\cite{dong2019evading}, \cite{yao2019trust}, \cite{dong2019efficient}.
There are also instances of exploiting unusual natural properties of the objects for model fooling~\cite{alcorn2019strike}, \cite{zhao2017generating} and attacking deep learning beyond the image space~\cite{zeng2019adversarial}, including 3D point cloud domain~\cite{xiang2019generating}, human skeletons~\cite{Liu2019AdvAttack} and the real-world~\cite{athalye2017synthesizing}, \cite{rey2019physical}. 


For the aforementioned attacks, a perturbation is estimated for a single input sample. Moosavi-Dezfooli et al.~\cite{moosavi2017universal} computed an input perturbation to fool a model into misclassifying `any' image with high probability. Similar `universal' adversarial perturbations also appear in \cite{khrulkov2018art}, \cite{mopuri2017fast}. These attacks are non-targeted, i.e.,~the adversarial input is allowed to be misclassified into any class. The universal  perturbations are able to reveal  interesting geometric correlations among the decision boundaries of the deep models \cite{moosavi2017universal}, \cite{moosavi2017analysis}. However,  these attacks neither restrict the input signals nor the predicted incorrect label for estimating the perturbations~\cite{moosavi2017universal}, \cite{khrulkov2018art}, \cite{mopuri2017fast}.
{\color{\ncolor}Hayes and Danezis~\cite{hayes2018learning} proposed Universal Adversarial Networks (UANs), which are generative models that can also compute targeted universal perturbations. Nevertheless, their technique does not consider restricting the scope of the input samples to a specific class, nor it suppresses the adversarial effects of the perturbations on irrelevant classes.
Manipulative signals maybe more revealing by systematically controlling the source and target classes and the effects of perturbations.
This partially motivates our adversarial attack that provides a comprehensive control over the perturbation computation with a transparent algorithm, which is not possible with 
any 
existing method.}


\vspace{-2mm}
\subsubsection{Adversarial defenses}
The ubiquitous prevalence of adversarial perturbations has naturally led to many defense techniques against the adversarial attacks~\cite{prakash2018deflecting}, \cite{akhtar2018defense}, \cite{raff2019barrage}, \cite{xie2019feature}, \cite{sun2019adversarial}, \cite{qiu2019adversarial}, \cite{jia2019comdefend}, \cite{liu2019detection}. However, adversarial attacks are often later found over-powering the  defenses~\cite{carlini2017adversarial}, \cite{athalye2018robustness}. We refer the interested readers to the living document maintained by Carlini et al.~\cite{carlini2019evaluating} for a thorough defense evaluation to avoid such scenarios. 
In general, the defense techniques aim at protecting deep models against both image-specific~\cite{prakash2018deflecting} and universal perturbations~\cite{akhtar2018defense}. This is commonly achieved by either detecting the perturbation in an input image, or diluting the adversarial effects of perturbation signals by modifying the model or the input itself. Nevertheless, the existence of a single unbreakable defense for all attacks still eludes the existing literature~\cite{carlini2017adversarial, carlini2016defensive, carlini2017magnet}, \cite{athalye2018robustness}.
Since our main focus is on adversarial attacks, we refer the interested readers to the  recent surveys~\cite{akhtar2018threat}, \cite{yuan2019adversarial} for more defense techniques. 

\vspace{-2mm}
\subsection{Non-adversarial perspective} Currently, there are also contributions in the literature, albeit very few, that portend the utility of perturbations beyond model fooling. For instance, Tsipras et al.~\cite{tsipras2019robustness} identified the presence of salient visual features of the target class in the perturbation signals that fool `adversarially robust' models, where the models have been robustified by including adversarial samples in the training data.  Woods et al.~\cite{woods2019reliable} also made a similar observation for the models robustified with regularized gradients. The existence of salient visual features in perturbations indicate the potential of these signals in model explanation~\cite{madry2017towards, woods2019reliable}. However, their manifestation \textit{uniquely} in the case of robustified models, is interpreted as a misalignment between (non-robust) deep representation and the human perception~\cite{engstrom2019learning, tsipras2019robustness}.
Potentially, the re-alignment is only achievable by making the models robust at a serious cost of performance loss and amplified computational complexity through adversarial training~\cite{engstrom2019learning, tsipras2019robustness}.  

It is worth indicating that the idea of  using input perturbation \textit{explicitly} for explanation/interpretation also exists in the literature. For instance, Fong et al.~\cite{fong2019understanding} proposed an `extremal' perturbation to identify the regions of an input image which strongly affect model predictions. A similar concept of `meaningful' perturbations was previously introduced by Fong and Vedaldi~\cite{fong2017interpretable}.
{\color{\ncolor}Other works that discuss the relation between perturbations and interpretation include~\cite{etmann2019connection},\cite{xu2018structured}, \cite{papernot2016limitations}, \cite{elliott2019adversarial}. However, those works either deal with input-specific interpretations~\cite{papernot2016limitations}, \cite{elliott2019adversarial}  that aim at highlighting the important image regions (similar to \cite{fong2019understanding},  \cite{fong2017interpretable}), or they deal with the interpretation of salient regions for input-specific  perturbations  themselves~\cite{etmann2019connection}, \cite{xu2018structured}. 
This is different from our model-centric view of interpretability, for which our attack generates input-agnostic visualizations. Our} perturbation is applied to the whole image, as normally done in the gradient-based adversarial attacks~\cite{goodfellow2014explaining}, \cite{madry2017towards}. {\color{\ncolor}Moreover,} we make the perturbation image-agnostic, thereby encoding the `model' information in the computed patterns instead of focusing on individual input samples.    

{\color{\ncolor}
Besides the perturbation based explanations, Ghorbani et al.~\cite{ghorbani2019towards} developed a framework to automatically identify human-understandable visual concepts deemed important by a model for its prediction. Their technique aggregates related segments across the inputs to identify high-level concepts for model explanation. Zhou et al.~\cite{zhou2018interpretable} decomposed the neural activations of an input into different semantic components that are subsequently used to generate human-meaningful input-specific interpretations. Previously,  \cite{yosinski2015understanding} also proposed to visualize network activations to interpret their decisions. 
They additionally proposed loss regularization to visualise the activation patterns in the input space preferred by the model. In contrast to \cite{yosinski2015understanding}, we achieve more clear visualisations without any such loss.  
In \cite{nguyen2017plug}, Nguyen et al.~introduced a Plug \& Play Generative Network to generate class-conditioned photo-realistic images with the help of class features extracted from a classifier. Such visualizations may be utilized in understanding the concepts learned by the classifier. Model-faithfulness of such images remains an open question though.     
Bau et al.~\cite{bau2018gan} specifically focused on studying the internal representation of GANs, proposing a method to visualise it with a segmentation-based method at different levels of abstraction for human-understanding. Though related to model interpretation, these methods do not deal with adversarial attacks on deep learning.  
}

\vspace{-2mm}
\section{Problem formulation}
\label{sec:PF}
Let $\boldsymbol{I} \in \mathbb R^m$ denote a sample of a distribution $\mathcal {I}$ over the natural images. Let   $\mathcal K (\boldsymbol{I})$ be a deep visual classifier  that maps $\boldsymbol{I}$ to its label $\ell_{\text{true}}$. In a typical adversarial setting, the objective is to generate a perturbation  $\boldsymbol{p} \in \mathbb R^m$ that satisfies the  constraint 
\begin{align}
    \mathcal K (\boldsymbol{I} + \boldsymbol{p}) \rightarrow \ell_{\text{target}}~~\text{s.t.}~ \ell_{\text{target}} \neq \ell_{\text{true}}, ||\boldsymbol{p}||_{p} \leq \eta, 
    \label{eq:pert}
\end{align}
where $||.||_{p}$ denotes the $\ell_p$-norm of the vector,  restrained by a pre-fixed  ${\eta}$. In (\ref{eq:pert}), restricting $\ell_{\text{target}}$ to a pre-defined label results in a targeted adversarial attack.

Based on (\ref{eq:pert}), it is possible to express $\boldsymbol{p}$ as a function over $\boldsymbol{I}$ and $\mathcal{K}(.)$\footnote{Assuming a fixed deterministic  algorithm to generate  $\boldsymbol{p}$.}. For a given  $\mathcal{K}(.)$, computing an image-specific perturbation confines the domain of $\boldsymbol{p}$, say $\text{Dom}(\boldsymbol{p})$ to a single image.
Implying, the information encoded in $\boldsymbol{p}$ is restricted to a single data sample. 
{\color{\ncolor} This does not make $\boldsymbol{p}$ a good indicator of the general character of the classifier.
This observation also weakens the argument of human perceptual misalignment with deep representation by alluding to image-specific perturbations.}
To truly encode classifier information in the perturbation, this signal needs to be invariant to the input samples, which we can achieve by broadening the domain of $\boldsymbol{p}$. 

Incidentally, universal adversarial perturbations~\cite{moosavi2017universal} are computed with a broader domain as per our formulation. 
Inline with our observation, those perturbations manifest  much more regular geometric patterns as compared to  image-specific perturbations. However, they still remain far from salient visual features of objects. This happens because universal perturbations map images to random class labels for the sake of model fooling. 
For a given classifier, broadening the perturbation domain with a `targeted' objective is more likely to induce geometric patterns in $\boldsymbol{p}$ that can actually be considered salient features of $\ell_{\text{target}}$, as viewed by the classifier. {\color{\ncolor} We also provide further evidence in this regard in \S~8a of the supplementary material of the paper.} 

Further to the above argument, we can alternately describe the objective of (\ref{eq:pert}) as maximizing the probability of a perturbed sample being mapped to $\ell_{\text{target}}$ by $\mathcal K(.)$. 
For $|\text{Dom}(\boldsymbol{p})|\gg 1$, where $|.|$ is the set cardinally, this maximization must incorporate all the relevant input samples (i.e.,~input image distribution). 
Ignoring $\ell_{\text{true}}$ of the input samples for a moment, we define the following broad objective based on (\ref{eq:pert}):
\begin{gather}
\label{eq:comb}
\max \underset{\text{Dom}(\boldsymbol{p})}{P}  \big( \mathcal K(\boldsymbol{I} + \boldsymbol{p})\rightarrow  \ell_{\text{target}} \big) \geq \gamma,~\text{s.t.} \\ \nonumber
||\boldsymbol{p}||_p \leq \eta, |\text{Dom}(\boldsymbol{p})| \gg \!1,
\end{gather}
where $P(.)$ denotes probability and $\gamma \in [0,1]$ is a pre-defined constant. 
To specify the `adversarial' and `explanation' usage of    perturbation under (\ref{eq:comb}), we define two separate sets of external constraints. For model fooling, we let
\begin{align*}
    \mathcal{C}_0^{f}:&~ \ell_{\text{target}} \neq \ell_{\text{true}},\\
    \mathcal{C}_1^{f}:& ~\text{Dom}(\boldsymbol{p}) = \{\boldsymbol{I} | \boldsymbol{I} \sim \mathcal I_{\text{source}} \},\\
    \mathcal{C}_2^f:&~ \underset{\text{Dom}^c(\boldsymbol{p})}{P}  \big( \mathcal K(\boldsymbol{I} + \boldsymbol{p})\rightarrow  \ell_{\text{target}} \big) < \gamma.
\end{align*}
Here, $\mathcal{C}_0^f$ ensures model fooling similar to the external constraint in (\ref{eq:pert}). $\mathcal{C}_1^f$ restricts the input domain to the image distribution $\mathcal{I}_{\text{source}}$ of a pre-defined source class. The constraint $\mathcal{C}_2^f$ mitigates the `leakage' of adversarial effects to the samples of non-source classes, specified  by $\text{Dom}^c(\boldsymbol{p})$.  

For the adversarial utility of our perturbation, it is more meaningful to have a stronger control over the input domain of this signal because fooling a  model on all inputs  alike (disregarding their correct labels) is hardly interesting. However, this is not the case for using the  perturbation for model explanation. For the later, image-agnostic nature of the task makes the label of input images largely irrelevant. Hence, the external constraint for the `explanation' objective is defined as follows:
\begin{align*}
    \mathcal{C}_1^e:&~ \text{Dom}(\boldsymbol{p}) = \{\boldsymbol{I} | \boldsymbol{I} \sim \mathcal{I} \} \cup \boldsymbol{d},
\end{align*}
where $\boldsymbol{d}$ is the `seed' sample that anchors the perturbations computation. Further details on `seed' are provided in \S~\ref{sec:ExpPComp} and \S~\ref{sec:ExpExplain}. 

\vspace{-2mm}
\section{Computing the perturbation}
\label{sec:Alg}
{\color{\ncolor} The primary contribution of this article is an attack method that is introduced with an adversarial objective (\S~\ref{sec:AdvP}), and then transformed into an attack with model explanation objective (\S~\ref{sec:ExpPComp}).}
For both `adversarial' and `explanation' objectives, our  perturbation estimation algorithms and the relevant  concepts are closely related. Hence, we first describe the algorithm fully with respect to the adversarial objective. Then, we specify the modifications to the algorithm to achieve the explanation objective.
{\color{\ncolor} The secondary contribution of utilizing the model explanation objective of our attack for low-level vision tasks is discussed in \S~\ref{sec:Evaluation} with empirical evidence.}

\vspace{-2mm}
\subsection{Adversarial perturbation computation}
\label{sec:AdvP}
We compute the perturbations for (targeted) fooling of a classifier as shown in  Algorithm~\ref{alg:main}. The algorithm optimizes for the objective in (\ref{eq:comb}), while satisfying the  constraints $\mathcal{C}_0^f$ to $\mathcal{C}_2^f$. The abstract concept of the   algorithm is intuitive. For a given source class, we compute the desired perturbation by taking small steps over the model's cost surface in the directions that increase the log-probability of the target label for the source class. The directions are computed stochastically, and the steps are only taken in the trusted regions that are governed by the first and (raw) second moment estimates of the directions. While computing a direction, we ensure that it also suppresses the prediction of non-source classes as the target label. To bound the perturbation norm, we keep projecting the accumulated signal to the $\ell_p$-ball of the desired norm at each iteration. The text below sequentially explains each step of the algorithm in detail.

Our attack is white-box in nature, hence the algorithm expects the target classifier as one of its inputs. It also requires a set $\mathcal S$ of the source class samples, and a set $\overline{\mathcal S}$ that contains samples of the non-source classes.
These sets are formed by sampling $\text{Dom}(\boldsymbol{p})$ and $\text{Dom}^c(\boldsymbol{p})$ respectively, mentioned in the previous section. Henceforth, we use the newly introduced sets as proxies for input domains for clarity in terms of implementation.  
Other input parameters include the desired $\ell_p$-norm `$\eta$' of the perturbation, target label `$\ell_{\text{target}}$', mini-batch size `$b$' for  the underlying stochastic optimization, and the desired fooling ratio  `$\gamma$' - defined as the percentage of the source class samples predicted as the target class instances.

We momentarily defer the discussion on  hyper-parameters `$\beta_1$' and `$\beta_2$' on \textit{line 1} of the algorithm. In a given iteration, we first construct the sets  $\mathcal S_s$ and $\mathcal{S}_o$ by randomly sampling the source and non-source classes, respectively. The cardinality of these sets is fixed to `$\frac{b}{2}$' to keep the mini-batch size to  `$b$' (\textit{line 3}). Each element of both sets is then perturbed with the current estimate of the perturbation - operation denoted by symbol  $\ominus$ on \textit{line 4}. The chosen symbol emphasizes that  $\boldsymbol{p}_t$ is subtracted in our algorithm from all the samples to perturb them. The `Clip(.)' function clips the perturbed  samples to its valid range, e.g.,~$[0, 255]$ in the  case of an 8-bit image representation.

\vspace{1mm}
\noindent{\bf Lemma 4.1:} \textit{For a classifier $\mathcal K(.)$ with cross-entropy cost $\mathcal J(\boldsymbol{\theta}, {\bf s}, \ell)$, the log-probability of an input sample ${\bf s}$ classified as `$\ell$' increases in the direction $-\frac{\nabla_{{\bf s}} \mathcal J(\boldsymbol{\theta}, {\bf s},  \ell)}{||\nabla_{{\bf s}} \mathcal J(\boldsymbol{\theta}, {\bf s},  \ell)||_{\infty}}$, where $\boldsymbol{\theta}$ denotes the model parameters\footnote{The model parameters remain fixed throughout, hence we ignore $\boldsymbol{\theta}$ in  Algorithm~\ref{alg:main} and the related text.}.}\\
\noindent{\bf Proof:} We can write $\mathcal J(\boldsymbol{\theta}, {\bf s}, \ell) = - \log\left(\text{P}(\ell| {\bf s})\right)$ for $\mathcal K(.)$. Linearizing the cost and inverting the sign, the log-probability maximizes along  $\boldsymbol{\gamma} =  -\nabla_{{\bf s}} \mathcal J(\boldsymbol{\theta}, {\bf s}, \ell)$. With $||\boldsymbol{\gamma}||_{\infty} = \max_i|\gamma_i|$, $\ell_{\infty}$-normalization re-scales $\boldsymbol{\gamma}$ in the same  direction of increasing $\log\left(\text{P}(\ell | {\bf s})\right)$.

\newcommand{\Exp}[1]{\underset{#1}{\mathbb E}}
\begin{algorithm}[t]
 \caption{Attack to fool}
 \label{alg:main}
 \begin{algorithmic}[1]
 \renewcommand{\algorithmicrequire}{\textbf{Input:}}
 \renewcommand{\algorithmicensure}{\textbf{Output:}}
 \REQUIRE  Classifier $\mathcal K$, source class samples $\mathcal S$, non-source class samples $\overline{\mathcal{S}}$, target label $\ell_{\text{target}}$, perturbation norm $\eta$, mini-batch size $b$,  fooling ratio $\gamma$.
 \ENSURE Perturbation $\boldsymbol{p} \in \mathbb R^{\text{m}}$.
 \STATE Initialize $\boldsymbol{p}_0$, $\boldsymbol{\upsilon}_0$, $\boldsymbol{\omega}_0$ to zero vectors  in $\mathbb R^{\text{m}}$ and $t = 0$. Set  $\beta_1 = 0.9$, and $\beta_2 = 0.999$.
\WHILE {fooling ratio $< \gamma$}  
\STATE $\mathcal S_s \isEq{\text{rand}} \mathcal S$,~$\mathcal S_o \isEq{\text{rand}} \overline{\mathcal S}$ : $|\mathcal S_s| = |\mathcal S_o| = \frac{b}{2}$  
\STATE ${\mathcal S_s} \leftarrow \text{Clip} \left ( \mathcal S_s \ominus \boldsymbol{p}_{t} \right)$,~$ {\mathcal S_o} \leftarrow \text{Clip} \left ( \mathcal S_o \ominus \boldsymbol{p}_{t} \right)$ 
\STATE $t \leftarrow  t+1$               
\STATE $\delta \leftarrow \frac{\Exp{{\bf s}_i \in \mathcal S_s}[|| \nabla_{{\bf s}_i} \mathcal J({\bf s}_i, \ell_{\text{target}})||_2 ]}  {\Exp{{\bf s}_i \in \mathcal S_o}[|| \nabla_{{\bf s}_i} \mathcal J({\bf s}_i, \ell)||_2 ]}$ 
\STATE $\boldsymbol\xi_t \leftarrow   \frac{1}{2}\Big(\Exp{{\bf s}_i \in \mathcal S_s} \big[ \nabla_{{\bf s}_i} \mathcal J({\bf s}_i, \ell_{\text{target}}) \big] +...$  \\
$~\hspace{37mm}\delta \Exp{{\bf s}_i \in \mathcal S_o} \big[ \nabla_{{\bf s}_i} \mathcal J({\bf s}_i, \ell) \big] \Big) $  
\STATE $\boldsymbol\upsilon_t \leftarrow \beta_1 \boldsymbol{\upsilon}_{t-1} + (1-\beta_1) \boldsymbol{\xi}_t$  
\STATE $\boldsymbol{\omega}_t \leftarrow \beta_2 \boldsymbol{\omega}_{t-1} + (1 - \beta_2) (\boldsymbol{\xi}_t\odot \boldsymbol{\xi}_t)$ 
\STATE $\boldsymbol{p} \leftarrow \frac{\sqrt{1-\beta_2^t}}{1-\beta_1^t}~\text{diag}\left( \text{diag}(\sqrt{\boldsymbol{\omega}_t})^{-1} \boldsymbol{\upsilon}_t \right)$ 
\STATE $\boldsymbol{p_t} \leftarrow \boldsymbol{p}_{t-1} +~\frac{\boldsymbol{p}}{|| \boldsymbol{p}||_{\infty}}$ 
\STATE $\boldsymbol{p_t} \leftarrow   \Psi (\boldsymbol{p}_t)$ 
\ENDWHILE
  \STATE return 
 \end{algorithmic}
 \end{algorithm}
\vspace{0.5mm}
 
Under Lemma 4.1, the algorithm strives to take steps along the cost function's gradient \textit{w.r.t.}~an  input ${\bf s}_i$. Since the domain of ${\bf s}_i$ spans multiple samples in our case, we must take steps along the `Expected' direction of those samples. However, we must ensure that the computed direction is not too generic to also cause the log-probability to rise for  irrelevant (i.e.~non-source class) samples. 
To refrain from the general fooling directions, we  nudge the computed direction such that it inhibits the fooling of non-source class samples. \textit{Lines 6 and 7} of the algorithm implement these steps to account for $\mathcal C_2^f$ as follows.  

On \textit{line 6}, we estimate the ratio between the Expected norms of the source sample gradients and the non-source sample gradients. Notice that we compute the respective gradients using different prediction labels. In the light of Lemma 4.1, $\nabla_{{\bf s}_i} \mathcal J({\bf s}_i, \ell_{\text{target}}): {\bf s}_i \in \mathcal S_s$ gives us the direction (ignoring the negative sign) to fool the model into predicting label `$\ell_{\text{target}}$' for ${\bf s}_i$, where the sample is from the source class. On the other hand, $\nabla_{{\bf s}_i} \mathcal J({\bf s}_i, \ell): {\bf s}_i \in \mathcal S_o$ provides the direction that improves the model confidence on the correct prediction of ${\bf s}_i$, where the sample is from the non-source class.
The diverse nature of the computed gradients can result in a significant difference between their norms. The scaling factor `$\delta$' on \textit{line 6} is computed  to account for that difference in the subsequent steps.
For the $t^{\text{th}}$ iteration, we compute the Expected gradient $\boldsymbol{\xi}_t$ of our mini-batch on \textit{line~7}.  At this point, it is worth noting that the effective mini-batch for the underlying stochastic optimization in our algorithm comprises clipped samples in the set $\mathcal S_s \bigcup \mathcal S_o$. The vector $\boldsymbol{\xi}_t$ is computed as the weighted  average of the Expected gradients of the source and non-source samples. Under the linearity of the Expectation operator and preservation of the vector direction with  scaling, it is straightforward to see that $\boldsymbol{\xi}_t$ encodes the Expected direction to achieve the targeted fooling of the source samples into the label `$\ell_{\text{target}}$', while inhibiting the fooling of non-source samples by increasing their prediction confidence for their correct classes.   

Owing to the diversity of the samples in its mini-batch, the algorithm steps in the direction of computed gradient  cautiously. 
On \textit{line 8} and \textit{line 9}, it respectively estimates the first and the raw second moment (i.e.,~un-centered variance) of the computed gradient using exponential moving averages. The hyper-parameters `$\beta_1$' and `$\beta_2$' decide the decay rates of these averages, whereas $\odot$ denotes the Hadamard product. The use of moving averages as the moment estimates in our algorithm is inspired by the Adam algorithm~\cite{kingma2014adam}, which efficiently performs stochastic optimization. However, instead of using the moving averages of gradients to update the parameters (i.e.~model weights) as in~\cite{kingma2014adam}, we compute those for the Expected gradient and capitalize on the directions for perturbation estimation. Nevertheless, due to the similar physical significance of the hyper-parameters  $\beta_1,~\beta_2 \in [0,1)$ in our algorithm and Adam,  the performance of both algorithms largely remains insensitive to small changes to the values of these parameters. Following~\cite{kingma2014adam}, we fix  $\beta_1 = 0.9, \beta_2 = 0.999$ (\textit{line 1}). We refer to~\cite{kingma2014adam} for further details on the choice of these values for the gradient based stochastic optimization. 

In our algorithm, the gradient moment estimates are exploited in stepping along the cost surface. Effectiveness of the moments as stepping guides for stochastic optimization is already well-established~\cite{kingma2014adam}. Briefly ignoring the expression for $\boldsymbol{p}$ on \textit{line~10} of the algorithm, we compute this guide as the ratio between the moment estimates    $\frac{\boldsymbol{\upsilon}_t}{\sqrt{\boldsymbol{\omega}_t}}$, where the square-root accounts for $\boldsymbol{\omega}_t$ representing the `second' moment. Note that, we slightly abuse the notation here as both values are vectors. On \textit{line~10}, we use the mathematically correct expression, where diag(.) converts a vector into a diagonal matrix, or a diagonal matrix into a vector, and the inverse is performed element-wise. Another improvement in \textit{line~10} is through the use of the  `bias-corrected' ratio of the moment estimates instead. Moving averages are known to get heavily biased at early iterations. This becomes a concern when the algorithm can benefit from well-estimated initial points. In our experiments (\S \ref{sec:AdvExp}), we use our algorithm in that manner. Hence, bias-correction is accounted for in our technique. We provide a detailed derivation to arrive at the expression on \textit{line~10} of Algorithm~\ref{alg:main} in \S A-1 of the supplementary material.  

Let us compactly write $\boldsymbol{p} = \frac{\widetilde{\boldsymbol{\upsilon}}_t}{\sqrt{\widetilde{\boldsymbol{\omega}}_t}}$, where 
%
%
$\sim$
%
%
indicates that the vectors are bias-corrected.
It is easy to see that for a large  second moment estimate $\widetilde{\boldsymbol{\omega}}$, $\boldsymbol{p}$ shrinks. This is desirable because we eventually take a step along $\boldsymbol{p}$, and a  smaller step is preferable along the components that have larger variance.
The perturbation update step on \textit{line 11} of the algorithm further restricts $\boldsymbol{p}$ to unit $\ell_{\infty}$-norm.
To an extent, this relates to computing the gradient's sign in FGSM~\cite{goodfellow2014explaining}. However, most coefficients of $\boldsymbol{p}$ get restricted to smaller values in our case instead of $\pm 1$. As a side remark, we note that simply computing the sign of $\boldsymbol{p}$ for perturbation update eventually nullifies the advantages of the second moment estimate due to the squared terms.   
The $\ell_{\infty}$ normalization is able to preserve the required direction in our case, while taking full advantage of the second moment estimate.  

{\color{\ncolor} Our adversarial attack in Algorithm~\ref{alg:main} not only forces model fooling on the source class samples, it also encourages correct predictions on the non-source class samples. On one hand, this provides a better control over the perturbation scope, on the other, it makes the attack more practical. From  the adversarial viewpoint, suspicions of an  attack can be minimized by exclusively manipulating a single class, instead of letting the model predict the same label for all the images. The over-generalization of the adversarial effects in the latter case  makes the attack less practical.}


\vspace{-2mm}
\subsubsection{Variants of adversarial perturbation}
\label{sec:Unbounded}
Notice that Algorithm~\ref{alg:main} accumulates the signals computed at each iteration.
To restrict the norm of the accumulated perturbation, $\ell_p$-ball projection is performed. The use of different types of balls results in different variants of the algorithm. For the  $\ell_{\infty}$-ball projection,  we implement $\Psi (\boldsymbol{p}_t) = \text{sign}(\boldsymbol p_t) \odot \text{min}\left(\text{abs}(\boldsymbol p_t), \eta\right)$ on \textit{line 12}. In the case of $\ell_2$-ball projection, we use $\Psi (\boldsymbol{p}_t) = \text{min}\left(1, \frac{\eta}{||\boldsymbol{p}_t||_2}\right) \boldsymbol p_t$. These  projections respectively bound the $\ell_{\infty}$ and $\ell_2$ norms of the perturbations. We bound these norms to reduce the perturbation  perceptibility, which is in line with the existing  literature taking an `adversarial' perspective on perturbations. Differently from the literature, we also employ a variant of our algorithm in which $\Psi (\boldsymbol{p}_t) = \mathbb I(\boldsymbol{p_t})$, where $\mathbb I(.)$ is the identity mapping. In contrast to the typical use of perturbations in adversarial attacks, we employ the  perturbations resulting from this `unbounded' variant to explore the classification regions of the target model without restricting the perturbation norm.
{\color{\ncolor}Due to our systematic control over the perturbation scope, this variant promises to reveal interesting information about the  classification regions of deep models - empirically verified in  \S \ref{sec:LUTAU}.} 

\vspace{-2mm}
\subsection{Perturbation computation for model explanation}
\label{sec:ExpPComp}
Similar to the adversarial perturbations, we compute perturbations to explain deep representation with a stochastic optimization scheme. {\color{\ncolor}The algorithm can be considered a modification of} Algorithm~\ref{alg:main}. Hence, here we only discuss the major differences between the two to avoid repetition. The explanation is kept at a higher level of abstraction for clarity, and only the most relevant concepts are discussed in detail. A complete step-by-step discussion of the algorithm is provided in \S A-2 of the supplementary material.

Recall that the objective for the explanation perturbation is given in (\ref{eq:comb}), followed by the external constraint $\mathcal{C}_1^e$ in~\S \ref{sec:PF}. 
To compute the perturbation, we deviate from  Algorithm~\ref{alg:main} in three major aspects. (\textbf{i}) Instead of simply estimating the perturbation, we additionally refine it. Details of perturbation refinement are provided in \S \ref{sec:refine}. (\textbf{ii})~The optimization is anchored by a `seed' image. From the implementation viewpoint, the use of a seed image can be  conceptualized as replacing the source class in Algorithm~\ref{alg:main} with a single image. However, since there is no constraint over the input class here, the seed as well as the `non-source' class samples (as per Algorithm~\ref{alg:main}) are all forced to be fooled into the same pre-selected target label $\ell_{\text{target}}$. 
The overall optimization is anchored with the seed image by weighting the model gradients \textit{w.r.t.}~seed differently as compared to the gradients for the other samples - analogous to \textit{line 7} of Algorithm~\ref{alg:main}. Further details are in the supplementary material. (\textbf{iii}) Before updating the intermediate perturbation in a given iteration, we conduct an additional  binary search. We invert the perturbation direction and evaluate if the inverted direction is equally good (or better) than the original direction for our objective in (\ref{eq:comb}). If so, we use the inverted perturbation instead of the original one in the subsequent iteration.
This step is mainly introduced to diversify the patterns encoded in the perturbation. This diversification is now preferred because we are more interested in  holistic patterns instead of the individual pixel values under the new objective.

For the convenience of our discussion, we refer to the scheme described above as Algorithm {\color{red}1a} in the text to follow. Algorithm {\color{red}1a} with detailed discussion is provided in the supplementary material. 
One aspect worth highlighting here is that 
whereas the $\ell_p$-norm of a perturbation is restricted in adversarial settings for \textit{imperceptibility}, this constraint plays a different role for our  novel application of perturbation.
In this case, by iterative back-projections on the $\ell_p$-ball, we amplify those geometric patterns in the perturbation that strongly influence $\mathcal{K(.)}$ to predict $\ell_{\text{target}}$ as the label of all the input samples.
With successive back-projections, we let the geometrically salient feature of $\ell_{\text{target}}$ to emerge in our perturbations (Fig.~\ref{fig:refine}) that we subsequently refine iteratively with the method discussed in \S~\ref{sec:refine}.

\vspace{-2mm}
\subsubsection{Perturbation refinement}
\label{sec:refine}
The holistic treatment of perturbation in Algorithm {\color{red}1a} results in an unrestricted spread of energy over the whole perturbation signal.
To achieve finer patterns we let the technique focus more on the relevant regions of input samples with an adaptive filtering mechanism summarized in Algorithm~\ref{alg:filteration}.  
A key property of this mechanism is that it upholds the  model fidelity of the perturbation by assuming no external priors.

\begin{algorithm}[t]
 \caption{Perturbation refinement }
 \label{alg:filteration} 
 \begin{algorithmic}[1]
 \renewcommand{\algorithmicrequire}{\textbf{Input:}}
 \renewcommand{\algorithmicensure}{\textbf{Output:}}
 \REQUIRE  Classifier $\mathcal K$, perturbation  $\boldsymbol{p} \in \mathbb R^{\text{m}}$
 \ENSURE Refined perturbation $\boldsymbol{p}$
 \STATE Initialize $\boldsymbol{f}$ to  $\boldsymbol{0} \in \mathbb R^{\text{m}}$ \\Set $\bar{\mathcal K} = $ convolutional base of  $\mathcal K$, scale factor $\lambda = 5$
 \STATE $\Omega \leftarrow \bar{\mathcal K}(\boldsymbol{p})$ : $\Omega \in \mathbb{R}^{H \times W \times C}$ 
 \STATE $\boldsymbol{a} \leftarrow \frac{1}{C} \sum_{n=1}^{\mathcal{\text{C}}} \Omega^n$
  \STATE $ \tau \leftarrow~$ $\Psi(\boldsymbol{a})$
  \STATE {\bf if} $\boldsymbol{a}(\text{x,y}) >  \tau~~\text{{\bf then}}~~\boldsymbol{a}(\text{x,y}) = \lambda~~\text{{\bf else}}~~ \boldsymbol{a}(\text{x,y})=\text{0}$ 
  \STATE $\boldsymbol{f} \leftarrow~$upsample ($\boldsymbol{a}$) : $\boldsymbol{f} \in \mathbb R^{\text{m}}   $
  \STATE $\boldsymbol{p} \leftarrow \text{Clip}(\boldsymbol{p} \odot \boldsymbol{f}$)
  \STATE return 
 \end{algorithmic}
 \end{algorithm}

To refine the perturbation, the signal  is fed to the convolutional base $\bar{\mathcal K}(.)$ of the classifier (\textit{line 2}).
The output $\Omega$ of the base is a set of low resolution 2D signals, which are reduced to an average signal $\boldsymbol{a}$ on \textit{line 3}. This signal captures a rough silhouette of the salient regions in the input perturbation, which makes it a useful spatial filter for our technique. On \textit{line 4}, $\Psi(.)$ computes the Otsu threshold~\cite{otsu1979threshold} for the average signal, that is subsequently used to binarize the image on \textit{line 5}. We empirically set $\lambda = 5$ in this work.  The resulting image is up-sampled with bicubic interpolation~\cite{keys1981cubic} on \textit{line~6} to match the dimensions of the input perturbation $\boldsymbol{p}$. The scaled mask is applied to the perturbation, which is subsequently clipped to the valid dynamic range.

\begin{figure}[t]
    \centering
    \includegraphics[width=0.48\textwidth]{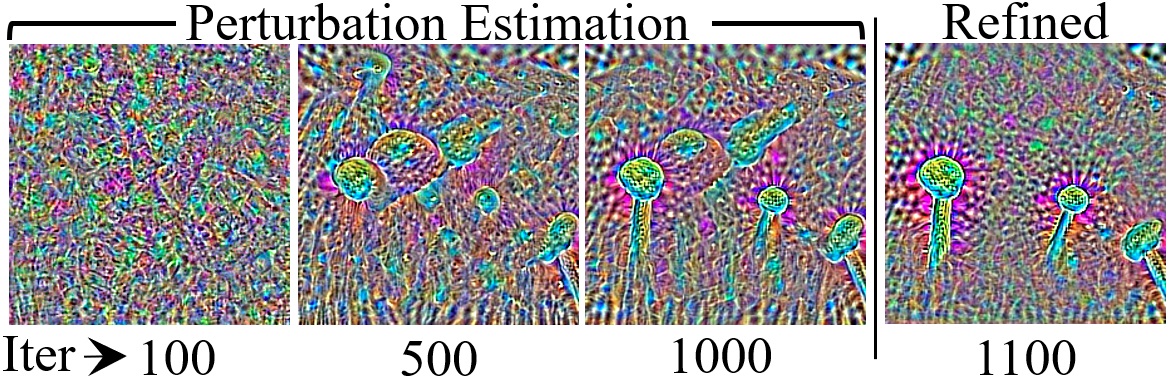}
    \caption{Visually salient geometric patterns emerge with more iteration of Algorithm~{\color{red}1a} (supplementary material) that are further refined with Algorithm~\ref{alg:filteration}. The refined perturbation is shown after post-refinement 100 iterations of the former. The `Nail' patterns are computed for VGG-16 with $\eta = 10$. We follow~\cite{szegedy2013intriguing} for perturbation visualization.}
    \label{fig:refine}
    \vspace{-3mm}
\end{figure}

The output of Algorithm~\ref{alg:filteration} is a refined perturbation that is again  processed by Algorithm~{\color{red}1a} to further highlight any salient patterns that might have been diminished with filtration. The final perturbation is computed by iterating between the two algorithms. In Fig.~\ref{fig:refine}, we show an example perturbation resulting from Algorithm {\color{red}1a} and after refinement by Algorithm~\ref{alg:filteration}.  

\vspace{-3mm}
\section{Evaluation}
\label{sec:Evaluation}
We thoroughly evaluate our perturbations from both the `adversarial' and `explanation' perspectives. In \S~\ref{sec:AdvExp}, we explore the model fooling efficacy of our  perturbations, including  a Physical World attack in \S~\ref{sec:Phy}, and also analyze the possibility of model exploration with unbounded adversarial perturbations in \S~\ref{sec:LUTAU}. We demonstrate the model explanation character of our perturbations in \S~\ref{sec:ExpExplain}, with a discussion on leveraging our perturbations for low level computational tasks in \S~\ref{sec:robust}.

\begin{table*}[t!]
\centering
{
\caption{{Fooling ratios (\%) with $\eta = 15$ for $\ell_{\infty}$ and $4,500$ for $\ell_2$-norm bounded  perturbations for ImageNet models.
The label transformations are T$_1$: Airship $\rightarrow$ School Bus, T$_2$: Ostrich $\rightarrow$ Zebra, T$_3$: Lion $\rightarrow$ Orangutang, T$_4$: Bustard $\rightarrow$ Camel, T$_5$: Jelly Fish $\rightarrow$ Killer Wahle, T$_6$: Life Boat $\rightarrow$ White Shark, T$_7$: Scoreboard $\rightarrow$ Freight Car, T$_8$: Pickelhaube $\rightarrow$ Stupa, T$_9$: Space Shuttle $\rightarrow$ Steam Locomotive, T$_{10}$: Rapeseed $\rightarrow$ Butterfly. 
{\color{\ncolorR}Average fooling ratio of the baseline - enhanced \cite{hayes2018learning}, is also provided in parentheses for reference.}
Leakage (last column) is the average fooling of non-source classes into the target label.}}
\vspace{-3mm}
\label{tab:success}
\begin{tabular}{c|l|c|c|c|c|c|c|c|c|c|c||c|c}
  \hline  
   {\bf Bound} & {\bf Model} & {\bf T$_1$} & {\bf T$_2$} & {\bf T$_3$} & {\bf T$_4$} & {\bf T$_5$} & {\bf T$_6$} & {\bf T$_7$} & {\bf T$_8$} & {\bf T$_9$} & {\bf T$_{10}$} &  {\bf Avg.} & {\bf Leak.} \\ \hline \hline
   \multirow{ 4}{*}{$\ell_{\infty}$-norm} & VGG-16~\cite{simonyan2014very} & 92  & 76  & 80  & 74  & 82  & 78  & 82  & 80 & 74  & 88  &  80.6$\pm$5.8 {\color{\ncolorR}(62.6$\pm$ 6.4)} & 29.9   \\
   & ResNet-50~\cite{he2016deep}     & 92 & 78  & 80  & 72  & 76  & 84  & 78  & 76  & 82  & 78  &  79.6$\pm$5.4 {\color{\ncolorR}(58.2$\pm$ 6.7)} &  31.1 \\
   & Inception-V3~\cite{szegedy2016rethinking} & 84 & 60  & 70  & 60  & 68  & 90  & 68  & 62  & 72  & 76 &  71.0$\pm$9.9 {\color{\ncolorR}(58.6$\pm$ 9.3)}  & 24.1 \\
   & MobileNet-V2~\cite{sandler2018mobilenetv2} & 92  & 94 & 88 & 78 & 88  & 86  & 74 & 86 & 84  & 94  &  86.4$\pm$6.5 {\color{\ncolorR}(64.6$\pm$ 5.9)} &  37.1 \\ \hline \hline
   \multirow{ 4}{*}{$\ell_{2}$-norm}      & VGG-16~\cite{simonyan2014very}           & 90  & 84  & 80  & 84  & 94  & 86  & 82  & 92  & 86  & 96  &  87.4$\pm$5.3 {\color{\ncolorR}(69.8$\pm$ 5.1)} & 30.4\\
   & ResNet-50~\cite{he2016deep}     & 96  & 94  & 88  & 84  & 90  & 86  & 86  & 94  & 90   & 90  &  89.8$\pm$3.9 {\color{\ncolorR}(70.4$\pm$ 4.5)} & 38.0\\
   & Inception-V3~\cite{szegedy2016rethinking} & 86  & 68  & 62  & 62 & 74  & 72  & 74  & 68  & 66  & 76  &  70.8$\pm$7.2 {\color{\ncolorR}(56.2$\pm$ 7.7)} & 45.6\\
   & MobileNet-V2~\cite{sandler2018mobilenetv2} & 94  & 98  & 92 & 76  & 94  & 92  & 76  & 92  &  92  & 96 &  90.2$\pm$7.7 {\color{\ncolorR}(73.4$\pm$ 7.3)} & 56.0\\ \hline

  \hline
\end{tabular}}
\end{table*}

\vspace{-3mm}
\subsection{Adversarial Experiments}
\label{sec:AdvExp}
\noindent{\bf Setup:} We first demonstrate the success of our targeted attack by fooling   VGG-16~\cite{simonyan2014very}, ResNet-50~\cite{he2016deep}, Inception-V3~\cite{szegedy2016rethinking} and MobileNet-V2~\cite{sandler2018mobilenetv2}  trained on ImageNet dataset~\cite{deng2009imagenet}. 
Our selection of the models is based on their established performance and diversity.  We use the training set of ILSVRC2012 for perturbation estimation, whereas the validation set of this data  (50 samples per class) is used as our test set.  For the non-source classes, we only use the correctly classified samples during training with a lower bound of $60\%$ on the prediction confidence. 
This filtration is  performed for computational purpose. It still ensures useful gradient directions  with fewer non-source samples. We do not filter the source class data. 
It is worth emphasizing that the existing works that compute universal perturbations, e.g.,~\cite{moosavi2017universal} only deal with the validation set of ImageNet. Comparatively, our data scope is orders of magnitude larger leading to more confident results.

We compute a perturbation using a \textit{two-step-strategy}. \textbf{First}, we alter Algorithm~\ref{alg:main} to disregard the non-source class data. This is achieved by  replacing the non-source class set $\overline{\mathcal S}$ with the source class set $\mathcal S$ and using `$\ell_{\text{target}}$' instead of `$\ell$' for the  gradient computation. In the \textbf{second} step, we initialize our algorithm with the perturbation computed in the first step.
This procedure is also adopted for computational gain with a better initialization. In the first step, we let the algorithm run for 100 iterations, while `$\gamma$' is set to $80\%$ in the second step. We additionally ensure at least 100 iterations in the second step. The batch size `$b$' is empirically set to 64 and 128  for the first and second step, respectively. In the text to follow, we discuss the setup details only when those are different from what is already described.     

Besides fooling the ImageNet models, we also attack the VGGFace model~\cite{cao2018vggface2} (ResNet-50 architecture) trained on the large-scale VGG-Face2 dataset~\cite{cao2018vggface2}. 
For that, 50 random images of an identity are used as the test set, while the remaining images are used for perturbation estimation. 

\vspace{1mm}
\noindent{\bf Fooling ImageNet models:}~We randomly choose ten source classes from  ImageNet and make another random selection of ten target labels, resulting in ten label transforming (i.e.,~fooling) experiments for a single model. Both $\ell_{\infty}$ and $\ell_2$-norm bounded perturbations are then  considered, letting  $\eta=15$ and $4,500$ respectively. As will be seen shortly, the perturbations remain imperceptible to quasi-imperceptible for these values. %

We summarize the results of our experiments in Table~\ref{tab:success}.  The reported fooling ratios are on test data that is previously unseen by both the targeted model and our algorithm. Successful fooling of the  models is apparent from the Table.
The table caption provides the label information for the source~$\rightarrow$~target transformation employing the commonly used nouns. We refer to \S A-3 of the supplementary material for the exact labels and original WordNet IDs of the ImageNet dataset. The last column reports the  `Leakage', which is defined as the average fooling ratio of the non-source classes into the target label. Hence, relatively low leakage is desirable, which is observed in the Table.  
{\color{\ncolor}
We explicitly analyse the success of leakage suppression in \S A-8 of the supplementary material}. 

{\color{\ncolorR} The table also provides baseline results for comparison, which is the average fooling ratio achieved by \cite{hayes2018learning} on the shown transformations. We only report the average value in Table~\ref{tab:success} in parentheses. For the complete results of \cite{hayes2018learning}, we refer to \S A-3 of the supplementary material.} 
{\color{\ncolor} We emphasize that the baseline results of the Universal Adversarial Network (UAN)~\cite{hayes2018learning} are achieved after considerable enhancement of the original UAN method. Since the adversarial objective of \cite{hayes2018learning} is easier than ours, a direct application of UAN to our problem resulted in only 10-15\% fooling ratios for the transformations considered in Table~\ref{tab:success}. To achieve comparable results, we enhanced UAN by utilizing the $\ell_p$-ball projection concept from our method. Specifically, we replaced the original process of monotonic perturbation norm increment after each iteration (which was also originally used to terminate the model training once a pre-define norm-threshold was achieved) with a projection step. Similar to our algorithm, the projection step allowed UAN to back-project  intermediate perturbations to a fixed  $\ell_p$-ball. This accumulated the intermediate perturbations more effectively. Since the UAN can not incorporate the leakage suppression that is possible with our method, we used the leakage as the termination criterion for the enhanced-UAN. For any transformation, we stopped the enhanced-UAN training when its leakage reached the corresponding leakage value for our method. Under a fair comparison that uses the same experimental setup for both methods, Table~\ref{tab:success} ascertains that our attack is still considerably stronger than UAN despite a significant enhancement of the latter.}

\begin{figure*}[t]
    \centering
    \includegraphics[height= 3.5in]{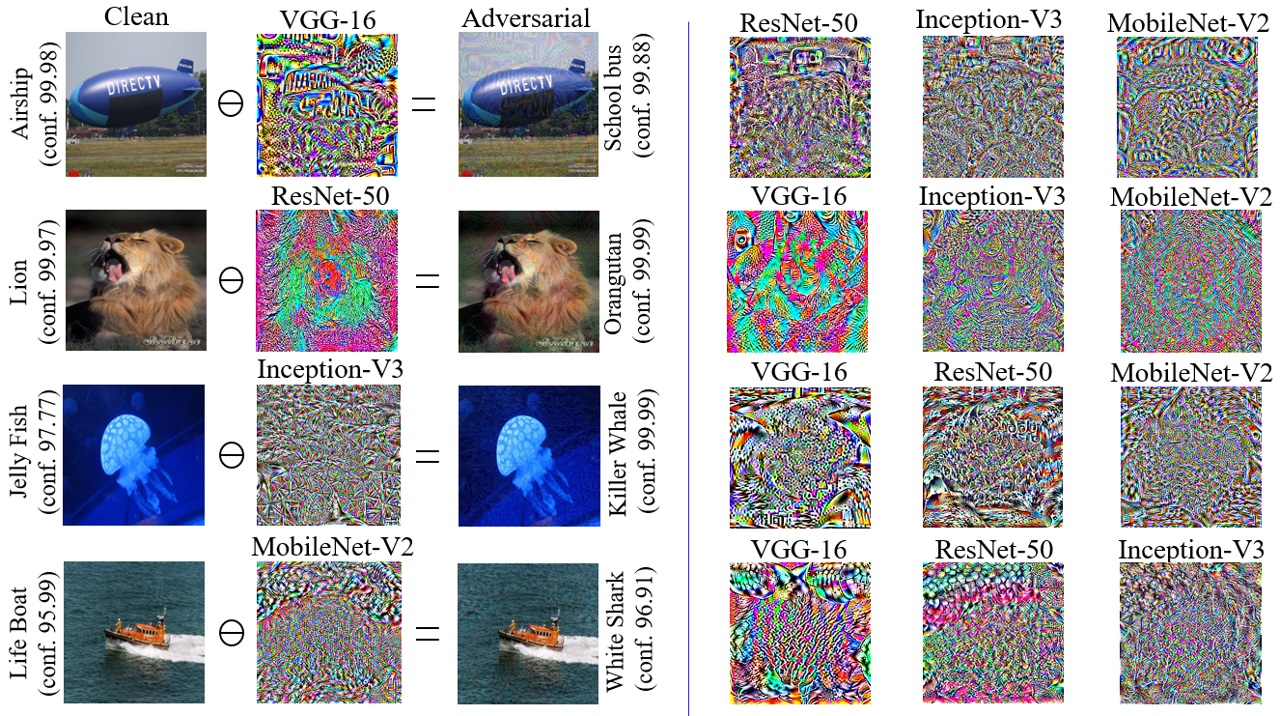}
    \caption{Representative  perturbations and adversarial images for $\ell_{\infty}$-bounded case ($\eta=15$). A row shows perturbations for the same source $\rightarrow$ target fooling of the mentioned models. An adversarial example for each model is also shown for reference (left). Following~\cite{szegedy2013intriguing},  the perturbations are magnified 10x, shifted by 128 and clamped to 0-255 for better visualization.}
    \label{fig:illus1}
    \vspace{-3mm}
\end{figure*}


In Fig.~\ref{fig:illus1}, we show   representative examples of label transformations. The figure includes a sample adversarial example for each network. In our experiments, it was frequently observed that the models show high confidence on the adversarial examples,
as stated in the figure.
We provide further images for both $\ell_{\infty}$ and $\ell_2$-norm perturbations in \S A-4 of the supplementary material. From the  images, we can observe that the perturbations are generally not easy to perceive visually. 

\begin{figure*}[t]
   \centering
    \includegraphics[width=0.8\textwidth]{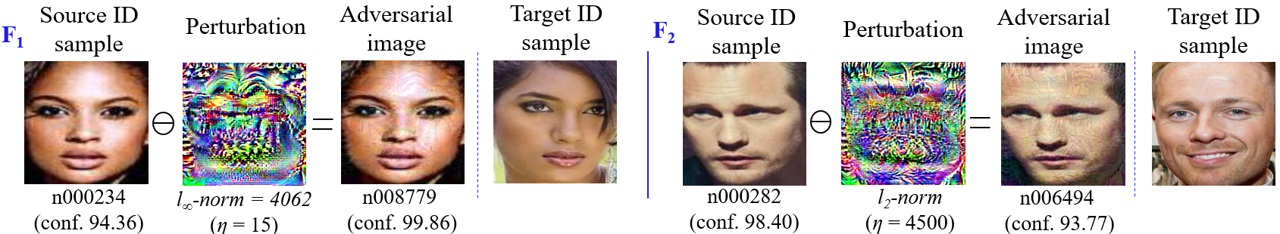}
    \caption{Representative face ID switching examples for VGGFace model. Sample clean target ID image is  provided for reference. }
    \label{fig:face}
\end{figure*}

\begin{table}[t]
       \caption{Switching face identities for VGGFace model (\% fooling): The switched identities in the original dataset are, F$_1$: n000234$\rightarrow$ n008779, F$_2$: n000282 $\rightarrow$ n006494, F$_3$: n000314 $\rightarrow$ n007087, F$_4$: n000558 $\rightarrow$ n001800,
    F$_5$: n005814 $\rightarrow$ n006402. The $\ell_{\infty}$ and $\ell_2$-norms of the perturbation are upper bounded to 15 and 4,500 respectively.}
    \centering
    \begin{tabular}{c|c|c|c|c|c|c}\hline
\multicolumn{7}{c}{$\ell_{\infty}$-norm bounded}\\
\hline \hline
  F$_1$ & F$_2$ & F$_3$ & F$_4$ & F$_5$ & Avg. & Leak. \\ \hline
  88 & 76 & 74 & 86 & 84 & 81.6$\pm$6.2 & 1.9\\ \hline
\multicolumn{7}{c}{$\ell_{2}$-norm bounded} \\
\hline \hline
    76 & 80 & 78 & 76 & 84 & 78.8$\pm$3.3 & 1.8 \\ \hline
    \end{tabular}
    \label{tab:face}
     \vspace{-3mm}
\end{table}

\vspace{1mm}
\noindent{\bf Fooling VGGFace model:}
Perhaps the most interesting application of our attack is in switching facial identities i.e. from a specific source identity to a specific target identity.
%
%
We demonstrate this by fooling the large-scale VGGFace model~\cite{cao2018vggface2}. 
Table~\ref{tab:face} reports the results on five identity switches that are randomly chosen from the  VGG-Face2 dataset. Considering the variety of  expression, appearance, ambient  conditions etc.~for a given subject in VGG-Face2, the results in Table~\ref{tab:face} imply that our perturbations enable an attacker to change their identity on-the-fly with high probability, without worrying about the image capturing conditions. Moreover, leakage of the  target label to the non-source classes also remains remarkably low for faces. 
Figure~\ref{fig:face} illustrates representative adversarial examples resulting from our algorithm for the face identity switches. 
Further images can also be found in \S A-5 of the supplementary material. 
The results ascertain successful identity switching on unseen inputs by our technique.

\begin{figure*}[t]
   \centering
\includegraphics[width=\textwidth, height = 1in]{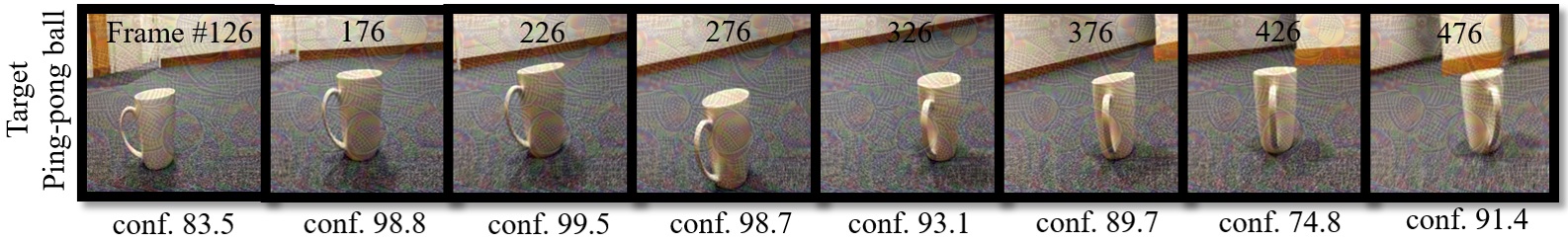}
\caption{Adding a single perturbation to a mobile camera video is able to consistently fool VGG-16 with high confidence (conf.) in real-time to misclassify a Coffee mug into a Ping-pong ball. Frames of a random clip are marked with their number.}
\label{fig:PingPong}
\end{figure*}

\subsubsection{Physical World attack}
\label{sec:Phy}
The proposed adversarial attacks have serious implications if they transfer well to the Physical World. To demonstrate this possibility, we select two pragmatic scenarios.   
In the first, we observe model predictions on a live webcam stream of printed adversarial images. No enhancement/transformation is applied to the images other than color printing. 
Due to geometric and illumination transformations in the Physical World, this setup is considerably more challenging than fooling classifiers on  digital scans of the printed adversarial images adopted in~\cite{kurakin2016adversarial}. Despite that, our perturbations are reasonably effective for targeted fooling in the Physical World. Successful fooling is observed even after slight rotations of the images. Further details of our experiments are provided in \S A-6 of the supplementary material and a video of live streaming is also included in the provided material.  

In the second scenario, we feed a single perturbation as an additive noise to a live mobile camera video to mimic a real-time attack. In Fig.~\ref{fig:PingPong}, we report the confidence of VGG-16 for classifying a Coffee Mug as a Ping-pong ball for every $50^{\text{th}}$ frame of a random clip in our video. A single perturbation computed by Algorithm~\ref{alg:main} is able to consistently fool the model on all the frames despite varied imaging conditions. This demonstrates the ability of our pre-computed perturbations to launch a real-time attack in practice.

\subsubsection{Explanation character of adversarial perturbations} 
\label{sec:LUTAU}
\begin{figure}[t]
  \begin{center}
    \includegraphics[width=0.45\textwidth]{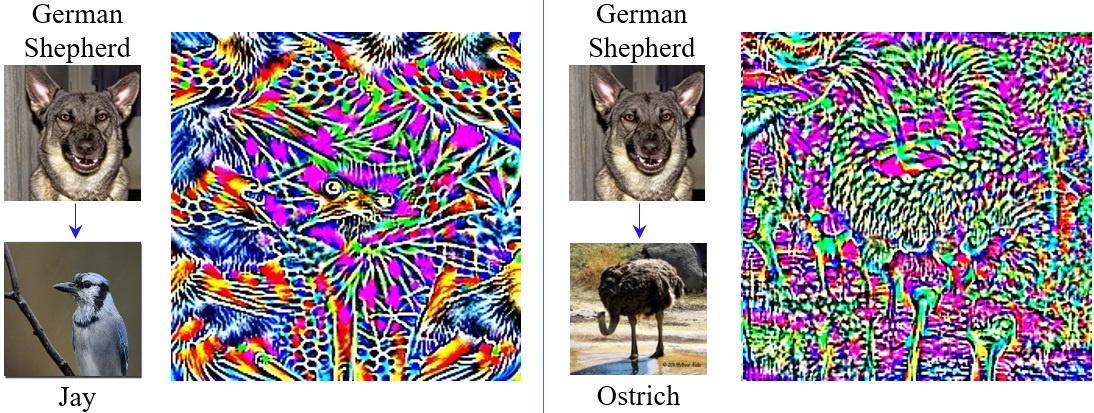}
  \end{center}
   \vspace{-3mm}
  \caption{Patterns resembling salient visual features of the target class emerge with perturbations allowed to achieve 100\% fooling rate on unseen data without norm restrictions. Representative perturbations for VGG-16 are shown for ImageNet validation set.}
  \label{fig:LUTAU-1}
\vspace{-3mm}
\end{figure}

Keeping aside the success of our targeted attack, we find the patterns fooling a deep model on a whole semantic concept interesting in their own right.
Hence, to investigate those, we let the unbounded variant of Algorithm~\ref{alg:main} discussed in \S~\ref{sec:Unbounded} run to achieve 100\% test accuracy and observe the computed  perturbation patterns. We notice a repetition of the salient visual features of the target class in the perturbations thus created, see  Fig.~\ref{fig:LUTAU-1}. 
We also observe that multiple runs of our algorithm resulted in different perturbations, nevertheless those perturbations preserve the characteristic features of the target label.
We refer  to \S A-7 of the supplementary material for the corroborating visualizations.
This observation advances the proposition that  perturbations with a broader input domain are inherently able to exploit the geometric correlations between the decision boundaries of the classifier~\cite{moosavi2017universal}.
{\color{\ncolor}This also entails that the human-understandable patterns emerging in the universal perturbations of \cite{poursaeed2018generative} may also be attributed to the target model instead of the generator network.}
In the supplementary material, we also provide perturbation patterns for the same target but different source classes. Those experiments confirm that the salient visual features of the target class become more pronounced in the perturbations with an increasing visual difference between the source and target classes.

\begin{table*}[t]
    \centering
    \caption{Average $\ell_2$-norms of the perturbations to achieve $95\%$ fooling on MobileNet-V2~\cite{sandler2018mobilenetv2}.}
         \vspace{-3mm}
    \label{tab:distance}
    \begin{tabular}{l||c|c|c|c|c} \hline
         {\bf Target} $\rightarrow$  & Space Shuttle & Steam Locomotive & Airship & School Bus & Life Boat\\
         {\bf Source} $\downarrow$ & &  & & &\\ \hline\hline
         Space Shuttle & - & 4364.4$\pm$81.1  & 4118.3$\pm$74.5 & 4679.4$\pm$179.5 & 5039.1$\pm$230.7 \\
         Steam Locomotive & 5406.8$\pm$57.7 & - & 4954.7$\pm$56.5 & 5845.2$\pm$300.4 & 5680.2$\pm$40.0 \\
         Airship & 3586.4$\pm$59.4 & 3992.7$\pm$291.1 & - & 3929.5$\pm$50.4 & 3937.8$\pm$33.7 \\
         School Bus & 7448.8$\pm$200.9 & 6322.8$\pm$89.5 & 6586.8$\pm$165.1 & - & 5976.5$\pm$112.1 \\
         Life Boat & 5290.4$\pm$43.1 & 5173.0$\pm$71.8 & 5121.5$\pm$154.1 & 5690.9$\pm$47.4 & - \\ \hline
    \end{tabular}
\end{table*}

We also explore another  interesting utility of our unbounded perturbations in terms of  exploring the classification regions induced by the deep models. 
We employ MobileNet-V2~\cite{sandler2018mobilenetv2}, and let the unbounded perturbations achieve $95\%$ fooling rate on the training samples in each experiment. The bound on fooling rate is for computational reasons only. We choose five ImageNet classes from Table~\ref{tab:success} and convert their labels into each other. We keep the number of training samples the same for each class, i.e.,~965 as allowed by the dataset. In our experiment, the perturbation vector's $\ell_2$-norm is used as the representative distance covered by the source class samples to cross over and stay in the  target class region. Experiments are repeated three times and the mean distances are reported in Table~\ref{tab:distance}. Interestingly, the differences between the distances for $A \rightarrow B$ and $B \rightarrow A$ are significant.
On the other hand, we can see  particularly lower values for `Airship' and larger values for `School Bus' for all transformations. These observations are explainable under the hypothesis that $w.r.t.$~the remaining classes, the classification region for `Airship' is more like a blob in the high dimensional space that lets the majority of the samples in it move (due to perturbations) more coherently towards other class regions. On the other end, `School Bus' occupies a relatively flat but well-spread  region that is farther from `Space Shuttle' as compared to e.g.,~`Life Boat'.

\begin{figure}[t]
    \centering
    \includegraphics[width = 0.49\textwidth]{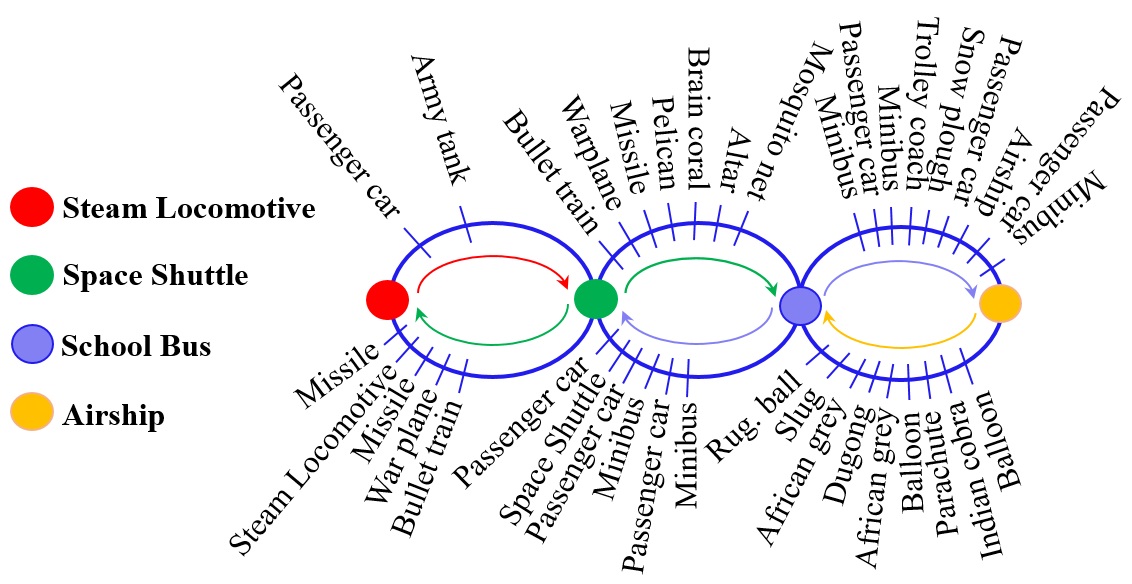}
    \caption{Max-label hopping during transformations with proposed unbounded adversarial perturbations. Setup of Table~\ref{tab:distance} is employed for transforming the four labels.}
    \label{fig:hops}
    \vspace{-3mm}
\end{figure}

Our algorithm makes the source class samples collectively move towards the target class region of a model with perturbations.
Hence, its iterations also provide a unique opportunity to examine this migration through the classification regions. For Table~\ref{tab:distance} experiment, we monitor the top-1 predictions during the iterations and record the \textit{maximally predicted labels} (excluding the source label) during training. In Fig.~\ref{fig:hops}, we show this information as `max-label hopping' for six representative transformations. 
Acute observers will notice that both Table~\ref{tab:distance} and Fig.~\ref{fig:hops} consider `transportation means' as the source and target classes.
This is done intentionally to illustrate the clustering of model  classification regions for semantically similar classes. 
Notice in Fig.~\ref{fig:hops} that the  hopping mostly involves intermediate classes related to transportation/carriage means. Exceptions occur when `School Bus' is the target class. This is inline with our hypothesis that this class has a well-spread region which allows it to attract a variety of intermediate labels as the target when perturbed, including those that live (relatively) far from its main cluster of transportation objects.

The above results  demonstrate an innate capacity of our 
adversarial perturbations to explore and explain deep models. In the Sections to follow, we verify this potential under the explanation variation of our perturbations, as explained in \S~\ref{sec:ExpPComp}.

\vspace{-2mm}
\subsection{Explanation Experiments}
\label{sec:ExpExplain}
Preliminary results of model {\color{\ncolor}interpretation through our attack to explain deep representation} were presented in CVPR2020~\cite{ourCVPR}.



\vspace{1mm}
\noindent{\bf Setup}: Our setup for these experiments is largely similar to the experiments for `Fooling ImageNet models' in \S~\ref{sec:AdvExp}. 
However, owing to the modified  Algorithm~{\color{red}1a} (supplementary material), we use a single seed image in place of the source class samples. 
We randomly choose the seed  and another set of 255 samples from the ImageNet validation set to emulate the non-source class samples. 
For all the 256 images, we never sample the target class, i.e.~$\ell_{\text{target}}$.
We set  $\gamma = 0.8$ and let $\eta = 10$. The value of `$\gamma$' is empirically chosen based on acceptable visual clarity of the salient patterns in the final perturbations. Larger values of this hyper-parameter generally leads to clearer patterns at the cost of higher computation. We keep  `$\eta$' comparable to the existing techniques for adversarial perturbation generation~\cite{moosavi2017universal, akhtar2018defense}. 
To compute a perturbation, we first let Algorithm~{\color{red}1a} run to achieve its stopping criterion. Then, we apply Algorithm~\ref{alg:filteration} for refinement. Subsequently, Algorithm~{\color{red}1a} is again applied such that a refinement is carried out after every $50^{\text{th}}$ iteration until $300$ iterations. 


\begin{figure*}[t]
    \centering
    \includegraphics[width=0.99\textwidth]{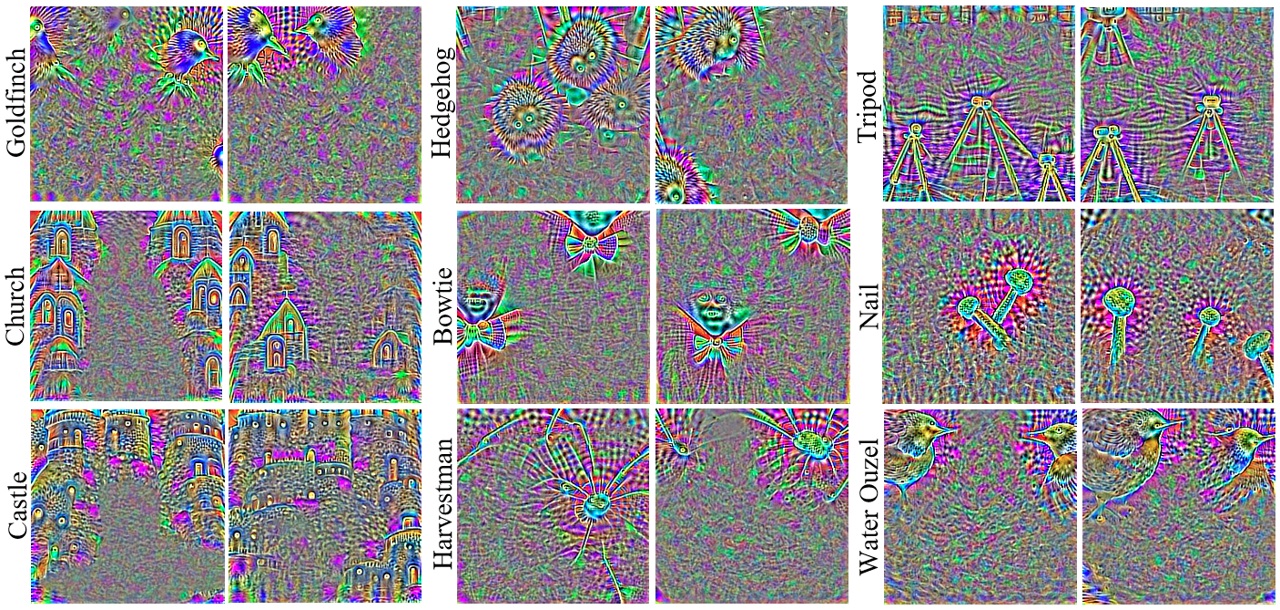}
    \caption{Well-localized visually salient features of the target class (label given) emerge by accumulating the gradient based perturbations with the explanation objective. The shown perturbations are computed for VGG-16 with ImageNet samples, excluding the target class samples. Perturbations for the same target are generated with different seeds for variety. Further images are also provided in \S A-9 of  supplementary material.}
    \label{fig:nonRobustIllustration}
\end{figure*}


\vspace{1mm}
\noindent{\bf Salient visual features}:
In Fig.~\ref{fig:nonRobustIllustration}, we show representative examples of the perturbations computed by our technique  for VGG-16. The target labels are also given for the shown two explanation perturbations for  each class.
Notice the well-localized clear geometric patterns that humans can associate with the target class labels. These patterns emerge without assuming any priors on the perturbation, input image distribution, or the model itself. Firstly, from the figure, it is apparent that our technique can (qualitatively) explain a model in terms of `what human-meaningful semantics are attached to its output neurons?'. This is useful e.g.,~in the settings where an unknown model is available and one must discover the labels of its output layer. Secondly, the perturbations are explaining `what geometric patterns are perceived as the discriminative features of a given class \textit{by the classifier}?'. Interestingly, these patterns align very well with the human perception, and we compute them with the same tool (i.e.,~gradient based perturbation) that is used to promote the argument of misalignment between human perception and deep representation~\cite{engstrom2019learning, tsipras2019robustness}. 

\vspace{1mm}
\noindent{\bf Diversity of the salient patterns}: We provide two representative perturbations for each target class in Fig.~\ref{fig:nonRobustIllustration}, where the difference in the perturbations is caused by selecting different seeds. Besides ascertaining the effective role of seed in our algorithm, the diverse patterns that remain visually salient,  affirm that the model has learned the general (human-meaningful) semantics for the target label. We emphasize that we have not used any sample of the target class in computing the respective perturbations in  Fig.~\ref{fig:nonRobustIllustration}.
The patterns are completely based on the visual model. This also highlights the potential of standard `classifiers' as `generators' of diverse images.


\vspace{1mm}
\noindent{\bf Region specific semantics} {\color{\ncolor}{\bf \& Patterns for different models}: 
We also successfully explore the possibility of analyzing the model semantics associated with specific image regions in \S A-9 of the supplementary material, where we additionally demonstrate that the manifestation of meaningful patterns in our perturbations is a generic phenomenon across the visual models.}

\begin{figure*}[t!]
    \centering
    \includegraphics[width=0.9\textwidth]{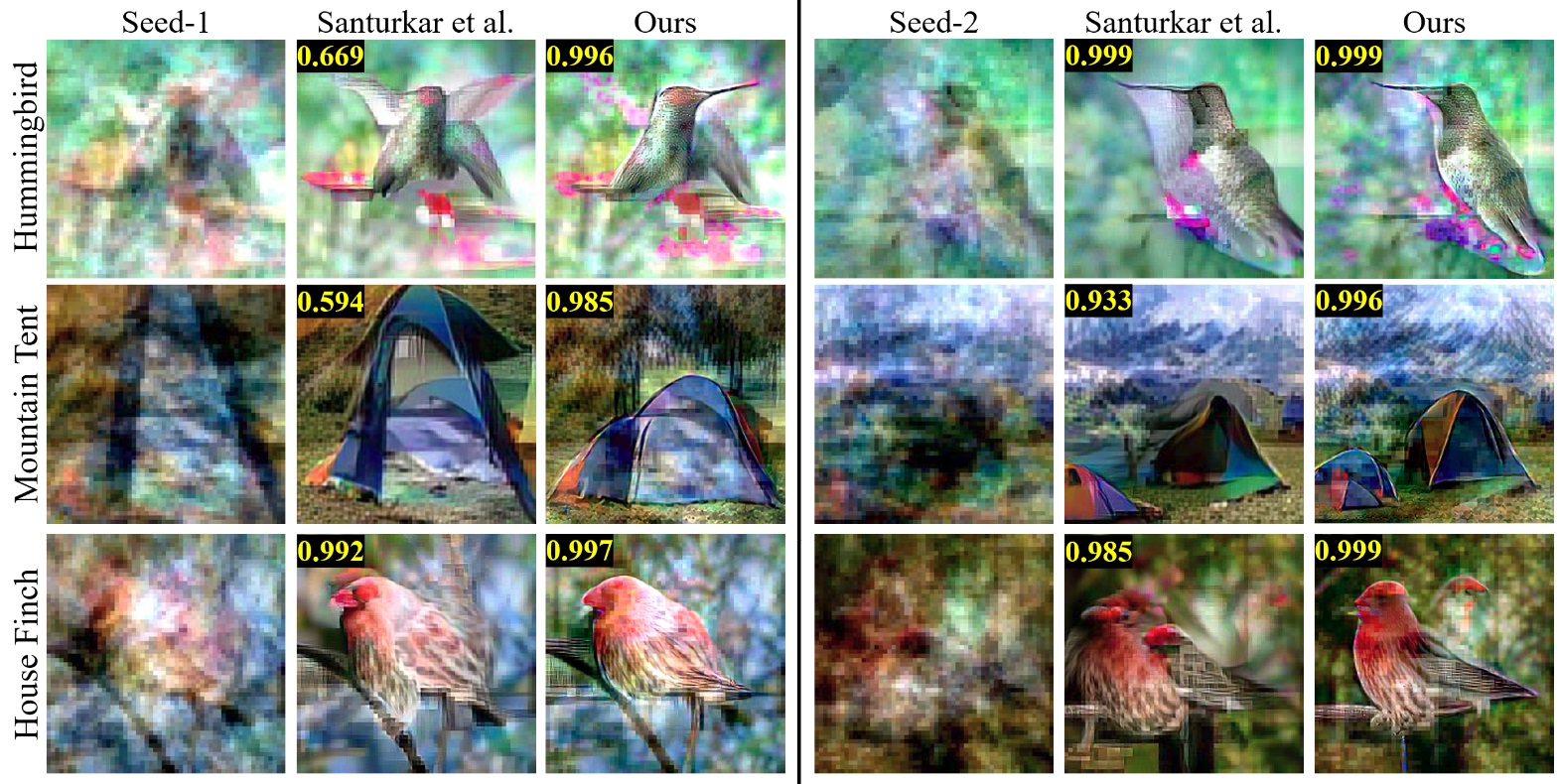}
    \caption{Image generation by attacking (adversarially) robust ResNet. The generated images are adversarial examples of the shown seeds. The intended class labels are mentioned on the left. We follow the setup of Santurkal et al.~\cite{santurkar2019computer} to generate images with both techniques. See \S A-11 of the supplementary material for further images. {\color{\ncolor} The confidence score of the robust classifier on the generated images is also reported.}}
    \label{fig:robustImageGeneration}
    \vspace{-3mm}
\end{figure*}


To further demonstrate the perceptual alignment of deep representation across different models, we also classify the  `perturbations' generated for one model with other models. High confidence of multiple models for the intended target label indicates that the extracted patterns are commonly seen as discriminative visual features of the target class. 
Details of our experiments are given in \S A-10 of the  supplementary material.
{\color{\ncolor} We note that in the above-mentioned tasks, a key requirement is the availability of clear information on the global structure and texture of the image. The shown results extract this information from the used robust models, which are known to encode it more clearly as compared to the  non-robust model~\cite{zhang2019interpreting}. Our algorithms  contribute towards refining this information by including even more interpretable patterns. To perform similar low-level vision tasks with non-robust models using our method, a more sophisticated external mechanism will be required, which is a future research direction for this work}.

\subsubsection{Leverage in low-level vision tasks}
\label{sec:robust}
Santurakar et al.~\cite{santurkar2019computer} recently showed that (adversarially) robust deep classifiers can be exploited for computer vision tasks beyond classification.
They demonstrated image generation, inpainting and image manipulation etc.~by attacking a robust ResNet  with the PGD attack~\cite{madry2017towards}.
The key concept on which Santurakar et al.~capitalized on is the presence of salient visual features in the adversarial perturbations computed with robust classifiers. 
Synchronizing this notion with our findings, their study identifies an ideal test-bed for our explanation attack. Successful results for these tasks by our attack cannot only ascertain the implicit model explaining nature of our perturbations, but can also improve the state-of-the-art for the newly found place of the robust classifiers in the broader Machine Learning context. 
{\color{\ncolor}Note that, the following experiments are not presented to suggest our method as an alternate to the state-of-the-art methods for the specified low-level vision tasks. Instead, the results are shown to demonstrate the explainability character of our method and further the notion of non-adversarial utilities of attack methods.}


To demonstrate improvements in the results, we follow~\cite{santurkar2019computer} closely  in terms of the used classifier, perturbation budget and the underlying evaluation procedure. In the experiments to follow, we create the input images for our algorithm by sampling a multivariate Gaussian $\mathcal N(\boldsymbol{\mu}_{I}, \Sigma_{I})$, where $\boldsymbol{\mu}_{I} \in \mathbb R^m$ is the mean value of an image set $\boldsymbol{I}_{i = 1,...,n}\!\!\sim\!\! \mathcal I^{\text{target}}$. Here,  $\mathcal I^{\text{target}}$ is the distribution of a target class images, emulated by ImageNet. We compute $\Sigma_I = \mathbb E[(\boldsymbol{I}_i - \boldsymbol{\mu}_{I})^{\intercal} (\boldsymbol{I}_i - \boldsymbol{\mu}_{I})]$.  
For computational reasons,  the multivariate Gaussian is computed by $4\times$ downsampling of the original images. Random $256$ distribution samples are later upsampled to match the network input and used to create the input images. Out of these images, one random image serves as the seed in an experiment.  In the experiments to follow, we do not use the refinement step (i.e.,~Algorithm~\ref{alg:filteration}) where the image processing tasks are performed holistically. 



\begin{figure*}[t]
    \centering
    \includegraphics[width=\textwidth]{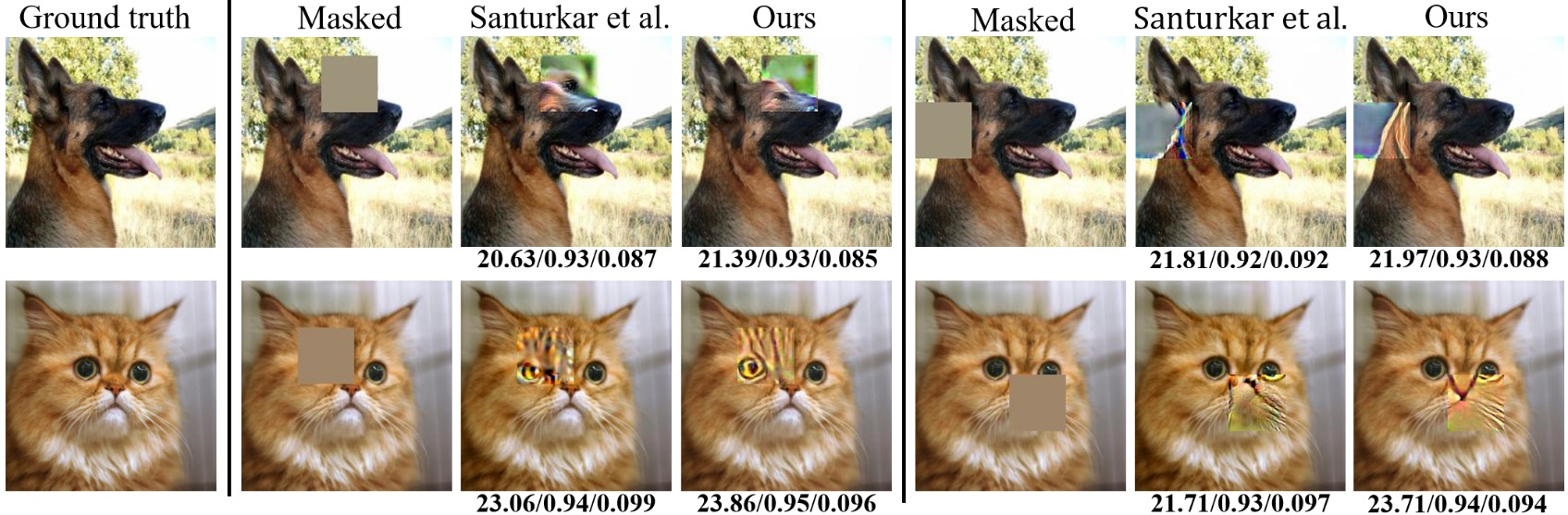}
    \caption{Representative inpainting results. The Masked image is used as the seed. Both methods restore images using the same robust model provided by Santurkar et al.~\cite{santurkar2019computer}, with the same perturbation budget. See \S A-12 of the supplementary material for more images. {\color{\ncolor} For each output image, PSNR/SSIM{\color{\ncolorR}/LPIPS} values are also provided.} {\color{\ncolorR} For LPIPS, lower values are more desirable.}}
    \label{fig:robustInpainting}
    \vspace{-2mm}
\end{figure*}

\noindent{\bf Image Generation:}
In Fig.~\ref{fig:robustImageGeneration}, we show representative examples of images generated by our technique and compare those with the results of Santurkal et al.~\cite{santurkar2019computer}. We use the author-provided code for \cite{santurkar2019computer} and strictly follow the guidelines to achieve the best results of their method. In the context of adversarial attacks, the generated images are adversarial examples of the seed images. We show two images per class, generated with the shown seeds for the mentioned target label. Our technique is  clearly able to generate more refined and coherent images. Notice the details in the backgrounds as well.
{\color{\ncolor} The figure also reports the confidence score of the robust classifier on the generated images. Considering that robust classifiers are able to focus more on the robust object features, higher confidence identifies better image generation.} Theoretically, Santurkar et al.~\cite{santurkar2019computer} used the strongest gradient-based iterative adversarial attack~\cite{madry2017towards} in their method. Hence, our improved performance can be clearly attributed to the model explaining nature of the perturbations computed by our technique. We use the same  perturbation budget $\eta = 40$ for both  techniques. {\color{\ncolorR} Overall, for the 1000 images we generated using the two methods, the average confidence score of the robust classifier on our images is $0.793\pm 0.171$, whereas this value for the corresponding images generated by \cite{santurkar2019computer} is $0.712\pm0.212$.}

The variety in the images generated with different seeds, their textural  details and clear semantic coherence strengthen the broader concept  that robust classifiers are capable of more than simple classification~\cite{santurkar2019computer}. It is a venue worth exploring for the future research.

\vspace{1mm}
\noindent{\bf Inpainting:}
Image inpainting~\cite{bertalmio2000image} restores information in large corrupt regions of images while upholding the perceptual consistency. 
For this task, we treat the corrupted  image as the seed, where its corrupt region  is identified as a binary mask $F \in \{0,1\}^m$.  Let $\Im$ be a set containing the seed and samples form our above-mentioned multivariate Gaussian distribution $\mathcal{N}(.)$. Keeping the robust classifier parameters fixed, we minimize the following loss:
\begin{align}
	\mathcal L (\boldsymbol{p}) = \mathbb E\big[\mathcal J(\Im_p, \ell_{\text{target}}) + \beta~\big( \boldsymbol{p} \odot (1 - F)\big) \big], 
\end{align}
where $\Im_p = \Im \ominus \boldsymbol{p}$, $\mathcal{J}(.)$ is the cross-entropy loss of the classifier and $\beta = 10$  is an empirically chosen  scaling factor.
The designed loss function allows the perturbation signal to grow freely for the corrupt region while restricting it in the other regions. This setup is also inspired by Santurkar et al.~\cite{santurkar2019computer}.

In Fig.~\ref{fig:robustInpainting}, we show representative examples of corrupt  images restored with our technique and Santurkar et al.~\cite{santurkar2019computer} using the robust ResNet provided by the authors. We use the same perturbation budget $\eta = 21$ for both techniques. The restoration quality of our technique is visibly better. The shown images and mask placements are randomly selected.
{\color{\ncolorR} For 1000 randomly selected images from the validation set of ImageNet and placing random masks on those images, the average PSNR/SSIM/LIPIS values for our method are $22.31\pm2.09$/$0.87\pm0.07$/$0.188\pm0.079$, whereas these values are $20.13\pm2.53$/$0.79\pm0.11$/$0.191\pm0.093$ for the corresponding images of  Santurkar et al.~\cite{santurkar2019computer}.}

 \vspace{1mm}
 \noindent{\bf Interactive Image Manipulation:}
An interesting recent application of deep neural networks, especially Generative Adversarial Networks~\cite{chen2018sketchygan} is to turn crude sketches into realistic images. Santurkar et al.~\cite{santurkar2019computer} demonstrated the possibility of such interactive image manipulation by adversarially  attacking robust classifiers. We advance this concept by demonstrating that our alternate objective of model explanation is more suitable for the problem. 

\begin{figure}[t]
    \centering
    \includegraphics[width=0.45\textwidth]{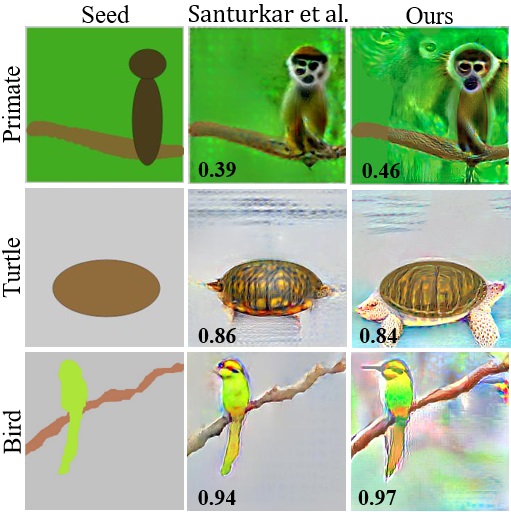}
    \caption{Representative examples of interactive image manipulation. The seed is a raw image required to be manipulated into an image of the target category. Both techniques use the same robust classifier with perturbation budget 60. See \S A-13 of the supplementary material for more visualizations. {\color{\ncolor} The confidence score of the robust classifier on the generated images is also provided.}}
    \label{fig:interactiveImageMan}
    \vspace{-4mm}
\end{figure}

Using the raw sketch as the seed and employing random samples from our multivariate Gaussian, we manipulate the seed similar to image generation. However, this time we also use the refinement process.  Representative results of our attack are shown in Fig.~\ref{fig:interactiveImageMan}. Compared to \cite{santurkar2019computer}, images generated with our attack appear to be much more realistic. Such refined manipulation of crude sketches with a classifier affirms the ability of our attack to highlight human-meaningful visual patterns learned by the classifier. {\color{\ncolorR} For the 100 random images we manipulated with our method, the average confidence score of the robust classifier is 0.79$\pm$0.108, whereas this value is 0.76$\pm$0.110 for the corresponding images manipulated by \cite{santurkar2019computer}}.  

\section{Conclusion}
{\color{\ncolor}We presented a novel attack for targeted fooling of deep visual models on individual object categories while suppressing the adversarial effects of perturbations on irrelevant classes. We modified the attack to also explain deep representation with input perturbations.}
To compute any perturbation, our attack performs a stochastic gradient search on the cost surface of the model to increase the  log-probability of a distribution of input images to be classified as a particular target. For fooling, it considers input samples which belong to the same class and suppresses any unintended fooling of any other class. We also show successful fooling results in the Physical World. For model explanation, by iterative back-projection of the gradients and refinement with adaptive attention, our attack finds geometric patterns in the perturbations that are deemed salient by the classifier. We find that these patterns align well with human perception, which weakens the argument of misalignment between human perception and deep representation - in the context of adversarial perturbations. We also demonstrate low-level image manipulation with our technique by attacking robust classifiers. With those experiments, we not only ascertain the model explaining nature of our attack by achieving realistic image generation, inpainting and interactive image manipulation results, we also advance the state-of-the-art of these newly found classifier utilities. 

\section{Acknowledgements}
This research was supported by ARC DP190102443 and the GPUs were donated by NVIDIA Corporation.


%







\balance
\bibliographystyle{IEEEtran}
\bibliography{main.bib}

\begin{thebibliography}{10}
\providecommand{\url}[1]{#1}
\csname url@samestyle\endcsname
\providecommand{\newblock}{\relax}
\providecommand{\bibinfo}[2]{#2}
\providecommand{\BIBentrySTDinterwordspacing}{\spaceskip=0pt\relax}
\providecommand{\BIBentryALTinterwordstretchfactor}{4}
\providecommand{\BIBentryALTinterwordspacing}{\spaceskip=\fontdimen2\font plus
\BIBentryALTinterwordstretchfactor\fontdimen3\font minus
  \fontdimen4\font\relax}
\providecommand{\BIBforeignlanguage}[2]{{%
\expandafter\ifx\csname l@#1\endcsname\relax
\typeout{** WARNING: IEEEtran.bst: No hyphenation pattern has been}%
\typeout{** loaded for the language `#1'. Using the pattern for}%
\typeout{** the default language instead.}%
\else
\language=\csname l@#1\endcsname
\fi
#2}}
\providecommand{\BIBdecl}{\relax}
\BIBdecl

\bibitem{szegedy2013intriguing}
C.~Szegedy, W.~Zaremba, I.~Sutskever, J.~Bruna, D.~Erhan, I.~Goodfellow, and
  R.~Fergus, ``Intriguing properties of neural networks,'' \emph{arXiv preprint
  arXiv:1312.6199}, 2013.

\bibitem{szegedy2016rethinking}
C.~Szegedy, V.~Vanhoucke, S.~Ioffe, J.~Shlens, and Z.~Wojna, ``Rethinking the
  inception architecture for computer vision,'' in \emph{Proceedings of the
  IEEE conference on computer vision and pattern recognition}, 2016, pp.
  2818--2826.

\bibitem{goodfellow2014explaining}
I.~J. Goodfellow, J.~Shlens, and C.~Szegedy, ``Explaining and harnessing
  adversarial examples,'' \emph{arXiv preprint arXiv:1412.6572}, 2014.

\bibitem{moosavi2016deepfool}
S.-M. Moosavi-Dezfooli, A.~Fawzi, and P.~Frossard, ``Deepfool: a simple and
  accurate method to fool deep neural networks,'' in \emph{Proceedings of the
  IEEE conference on computer vision and pattern recognition}, 2016, pp.
  2574--2582.

\bibitem{dong2018boosting}
Y.~Dong, F.~Liao, T.~Pang, H.~Su, J.~Zhu, X.~Hu, and J.~Li, ``Boosting
  adversarial attacks with momentum,'' in \emph{Proceedings of the IEEE
  conference on computer vision and pattern recognition}, 2018, pp. 9185--9193.

\bibitem{inkawhich2019feature}
N.~Inkawhich, W.~Wen, H.~H. Li, and Y.~Chen, ``Feature space perturbations
  yield more transferable adversarial examples,'' in \emph{Proceedings of the
  IEEE Conference on Computer Vision and Pattern Recognition}, 2019, pp.
  7066--7074.

\bibitem{moosavi2017universal}
S.-M. Moosavi-Dezfooli, A.~Fawzi, O.~Fawzi, and P.~Frossard, ``Universal
  adversarial perturbations,'' in \emph{Proceedings of the IEEE conference on
  computer vision and pattern recognition}, 2017, pp. 1765--1773.

\bibitem{li2019universal}
J.~Li, R.~Ji, H.~Liu, X.~Hong, Y.~Gao, and Q.~Tian, ``Universal perturbation
  attack against image retrieval,'' in \emph{Proceedings of the IEEE
  International conference on computer vision}, 2019.

\bibitem{akhtar2018threat}
N.~Akhtar and A.~Mian, ``Threat of adversarial attacks on deep learning in
  computer vision: A survey,'' \emph{IEEE Access}, vol.~6, pp.
  14\,410--14\,430, 2018.

\bibitem{ilyas2019adversarial}
A.~Ilyas, S.~Santurkar, D.~Tsipras, L.~Engstrom, B.~Tran, and A.~Madry,
  ``Adversarial examples are not bugs, they are features,'' \emph{arXiv
  preprint arXiv:1905.02175}, 2019.

\bibitem{deng2009imagenet}
J.~Deng, W.~Dong, R.~Socher, L.-J. Li, K.~Li, and L.~Fei-Fei, ``Imagenet: A
  large-scale hierarchical image database,'' in \emph{2009 IEEE conference on
  computer vision and pattern recognition}.\hskip 1em plus 0.5em minus
  0.4em\relax Ieee, 2009, pp. 248--255.

\bibitem{engstrom2019learning}
L.~Engstrom, A.~Ilyas, S.~Santurkar, D.~Tsipras, B.~Tran, and A.~Madry,
  ``Learning perceptually-aligned representations via adversarial robustness,''
  \emph{arXiv preprint arXiv:1906.00945}, 2019.

\bibitem{he2016deep}
K.~He, X.~Zhang, S.~Ren, and J.~Sun, ``Deep residual learning for image
  recognition,'' in \emph{Proceedings of the IEEE conference on computer vision
  and pattern recognition}, 2016, pp. 770--778.

\bibitem{simonyan2014very}
K.~Simonyan and A.~Zisserman, ``Very deep convolutional networks for
  large-scale image recognition,'' \emph{arXiv preprint arXiv:1409.1556}, 2014.

\bibitem{santurkar2019computer}
S.~Santurkar, D.~Tsipras, B.~Tran, A.~Ilyas, L.~Engstrom, and A.~Madry,
  ``Computer vision with a single (robust) classifier,'' \emph{arXiv preprint
  arXiv:1906.09453}, 2019.

\bibitem{sandler2018mobilenetv2}
M.~Sandler, A.~Howard, M.~Zhu, A.~Zhmoginov, and L.-C. Chen, ``Mobilenetv2:
  Inverted residuals and linear bottlenecks,'' in \emph{Proceedings of the IEEE
  Conference on Computer Vision and Pattern Recognition}, 2018, pp. 4510--4520.

\bibitem{cao2018vggface2}
Q.~Cao, L.~Shen, W.~Xie, O.~M. Parkhi, and A.~Zisserman, ``Vggface2: A dataset
  for recognising faces across pose and age,'' in \emph{2018 13th IEEE
  International Conference on Automatic Face \& Gesture Recognition (FG
  2018)}.\hskip 1em plus 0.5em minus 0.4em\relax IEEE, 2018, pp. 67--74.

\bibitem{ourCVPR}
M.~Jalwana, N.~Akhtar, M.~Bennamoun, and A.~Mian, ``Attack to explain deep
  representation,'' in \emph{Proceedings of the IEEE Conference on Computer
  Vision and Pattern Recognition}, 2020.

\bibitem{kurakin2016adversarial}
A.~Kurakin, I.~Goodfellow, and S.~Bengio, ``Adversarial examples in the
  physical world,'' \emph{arXiv preprint arXiv:1607.02533}, 2016.

\bibitem{wu2018understanding}
L.~Wu, Z.~Zhu, C.~Tai \emph{et~al.}, ``Understanding and enhancing the
  transferability of adversarial examples,'' \emph{arXiv preprint
  arXiv:1802.09707}, 2018.

\bibitem{xie2019improving}
C.~Xie, Z.~Zhang, Y.~Zhou, S.~Bai, J.~Wang, Z.~Ren, and A.~L. Yuille,
  ``Improving transferability of adversarial examples with input diversity,''
  in \emph{Proceedings of the IEEE Conference on Computer Vision and Pattern
  Recognition}, 2019, pp. 2730--2739.

\bibitem{madry2017towards}
A.~Madry, A.~Makelov, L.~Schmidt, D.~Tsipras, and A.~Vladu, ``Towards deep
  learning models resistant to adversarial attacks,'' \emph{arXiv preprint
  arXiv:1706.06083}, 2017.

\bibitem{shi2019curls}
Y.~Shi, S.~Wang, and Y.~Han, ``Curls \& whey: Boosting black-box adversarial
  attacks,'' \emph{arXiv preprint arXiv:1904.01160}, 2019.

\bibitem{rony2019decoupling}
J.~Rony, L.~G. Hafemann, L.~S. Oliveira, I.~B. Ayed, R.~Sabourin, and
  E.~Granger, ``Decoupling direction and norm for efficient gradient-based l2
  adversarial attacks and defenses,'' in \emph{Proceedings of the IEEE
  Conference on Computer Vision and Pattern Recognition}, 2019, pp. 4322--4330.

\bibitem{croce2019sparse}
F.~Croce and M.~Hein, ``Sparse and imperceivable adversarial attacks,'' in
  \emph{Proceedings of the IEEE International conference on computer vision},
  2019.

\bibitem{dong2019evading}
Y.~Dong, T.~Pang, H.~Su, and J.~Zhu, ``Evading defenses to transferable
  adversarial examples by translation-invariant attacks,'' in \emph{Proceedings
  of the IEEE Conference on Computer Vision and Pattern Recognition}, 2019, pp.
  4312--4321.

\bibitem{yao2019trust}
Z.~Yao, A.~Gholami, P.~Xu, K.~Keutzer, and M.~W. Mahoney, ``Trust region based
  adversarial attack on neural networks,'' in \emph{Proceedings of the IEEE
  Conference on Computer Vision and Pattern Recognition}, 2019, pp.
  11\,350--11\,359.

\bibitem{dong2019efficient}
Y.~Dong, H.~Su, B.~Wu, Z.~Li, W.~Liu, T.~Zhang, and J.~Zhu, ``Efficient
  decision-based black-box adversarial attacks on face recognition,'' in
  \emph{Proceedings of the IEEE Conference on Computer Vision and Pattern
  Recognition}, 2019, pp. 7714--7722.

\bibitem{alcorn2019strike}
M.~A. Alcorn, Q.~Li, Z.~Gong, C.~Wang, L.~Mai, W.-S. Ku, and A.~Nguyen,
  ``Strike (with) a pose: Neural networks are easily fooled by strange poses of
  familiar objects,'' in \emph{Proceedings of the IEEE Conference on Computer
  Vision and Pattern Recognition}, 2019, pp. 4845--4854.

\bibitem{zhao2017generating}
Z.~Zhao, D.~Dua, and S.~Singh, ``Generating natural adversarial examples,''
  \emph{arXiv preprint arXiv:1710.11342}, 2017.

\bibitem{zeng2019adversarial}
X.~Zeng, C.~Liu, Y.-S. Wang, W.~Qiu, L.~Xie, Y.-W. Tai, C.-K. Tang, and A.~L.
  Yuille, ``Adversarial attacks beyond the image space,'' in \emph{Proceedings
  of the IEEE Conference on Computer Vision and Pattern Recognition}, 2019, pp.
  4302--4311.

\bibitem{xiang2019generating}
C.~Xiang, C.~R. Qi, and B.~Li, ``Generating 3d adversarial point clouds,'' in
  \emph{Proceedings of the IEEE Conference on Computer Vision and Pattern
  Recognition}, 2019, pp. 9136--9144.

\bibitem{Liu2019AdvAttack}
J.~Liu, N.~Akhtar, and A.~Mian, ``Adversarial attack on skeleton-based human
  action recognition,'' \emph{arXiv preprint arXiv:1909.06500}, 2019.

\bibitem{athalye2017synthesizing}
A.~Athalye, L.~Engstrom, A.~Ilyas, and K.~Kwok, ``Synthesizing robust
  adversarial examples,'' \emph{arXiv preprint arXiv:1707.07397}, 2017.

\bibitem{rey2019physical}
R.~Wiyatno and A.~Xu, ``Physical adversarial textures that fool visual object
  tracking,'' in \emph{Proceedings of the IEEE International conference on
  computer vision}, 2019.

\bibitem{khrulkov2018art}
V.~Khrulkov and I.~Oseledets, ``Art of singular vectors and universal
  adversarial perturbations,'' in \emph{Proceedings of the IEEE Conference on
  Computer Vision and Pattern Recognition}, 2018, pp. 8562--8570.

\bibitem{mopuri2017fast}
K.~R. Mopuri, U.~Garg, and R.~V. Babu, ``Fast feature fool: A data independent
  approach to universal adversarial perturbations,'' \emph{arXiv preprint
  arXiv:1707.05572}, 2017.

\bibitem{moosavi2017analysis}
S.-M. Moosavi-Dezfooli, A.~Fawzi, O.~Fawzi, P.~Frossard, and S.~Soatto,
  ``Analysis of universal adversarial perturbations,'' \emph{arXiv preprint
  arXiv:1705.09554}, 2017.

\bibitem{hayes2018learning}
J.~Hayes and G.~Danezis, ``Learning universal adversarial perturbations with
  generative models,'' in \emph{2018 IEEE Security and Privacy Workshops
  (SPW)}.\hskip 1em plus 0.5em minus 0.4em\relax IEEE, 2018, pp. 43--49.

\bibitem{prakash2018deflecting}
A.~Prakash, N.~Moran, S.~Garber, A.~DiLillo, and J.~Storer, ``Deflecting
  adversarial attacks with pixel deflection,'' in \emph{Proceedings of the IEEE
  conference on computer vision and pattern recognition}, 2018, pp. 8571--8580.

\bibitem{akhtar2018defense}
N.~Akhtar, J.~Liu, and A.~Mian, ``Defense against universal adversarial
  perturbations,'' in \emph{Proceedings of the IEEE Conference on Computer
  Vision and Pattern Recognition}, 2018, pp. 3389--3398.

\bibitem{raff2019barrage}
E.~Raff, J.~Sylvester, S.~Forsyth, and M.~McLean, ``Barrage of random
  transforms for adversarially robust defense,'' in \emph{Proceedings of the
  IEEE Conference on Computer Vision and Pattern Recognition}, 2019, pp.
  6528--6537.

\bibitem{xie2019feature}
C.~Xie, Y.~Wu, L.~v.~d. Maaten, A.~L. Yuille, and K.~He, ``Feature denoising
  for improving adversarial robustness,'' in \emph{Proceedings of the IEEE
  Conference on Computer Vision and Pattern Recognition}, 2019, pp. 501--509.

\bibitem{sun2019adversarial}
B.~Sun, N.-h. Tsai, F.~Liu, R.~Yu, and H.~Su, ``Adversarial defense by
  stratified convolutional sparse coding,'' in \emph{Proceedings of the IEEE
  Conference on Computer Vision and Pattern Recognition}, 2019, pp.
  11\,447--11\,456.

\bibitem{qiu2019adversarial}
Y.~Qiu, J.~Leng, C.~Guo, Q.~Chen, C.~Li, M.~Guo, and Y.~Zhu, ``Adversarial
  defense through network profiling based path extraction,'' in
  \emph{Proceedings of the IEEE Conference on Computer Vision and Pattern
  Recognition}, 2019, pp. 4777--4786.

\bibitem{jia2019comdefend}
X.~Jia, X.~Wei, X.~Cao, and H.~Foroosh, ``Comdefend: An efficient image
  compression model to defend adversarial examples,'' in \emph{Proceedings of
  the IEEE Conference on Computer Vision and Pattern Recognition}, 2019, pp.
  6084--6092.

\bibitem{liu2019detection}
J.~Liu, W.~Zhang, Y.~Zhang, D.~Hou, Y.~Liu, H.~Zha, and N.~Yu, ``Detection
  based defense against adversarial examples from the steganalysis point of
  view,'' in \emph{Proceedings of the IEEE Conference on Computer Vision and
  Pattern Recognition}, 2019, pp. 4825--4834.

\bibitem{carlini2017adversarial}
N.~Carlini and D.~Wagner, ``Adversarial examples are not easily detected:
  Bypassing ten detection methods,'' in \emph{Proceedings of the 10th ACM
  Workshop on Artificial Intelligence and Security}.\hskip 1em plus 0.5em minus
  0.4em\relax ACM, 2017, pp. 3--14.

\bibitem{athalye2018robustness}
A.~Athalye and N.~Carlini, ``On the robustness of the cvpr 2018 white-box
  adversarial example defenses,'' \emph{arXiv preprint arXiv:1804.03286}, 2018.

\bibitem{carlini2019evaluating}
N.~Carlini, A.~Athalye, N.~Papernot, W.~Brendel, J.~Rauber, D.~Tsipras,
  I.~Goodfellow, and A.~Madry, ``On evaluating adversarial robustness,''
  \emph{arXiv preprint arXiv:1902.06705}, 2019.

\bibitem{carlini2016defensive}
N.~Carlini and D.~Wagner, ``Defensive distillation is not robust to adversarial
  examples,'' \emph{arXiv preprint arXiv:1607.04311}, 2016.

\bibitem{carlini2017magnet}
------, ``Magnet and" efficient defenses against adversarial attacks" are not
  robust to adversarial examples,'' \emph{arXiv preprint arXiv:1711.08478},
  2017.

\bibitem{yuan2019adversarial}
X.~Yuan, P.~He, Q.~Zhu, and X.~Li, ``Adversarial examples: Attacks and defenses
  for deep learning,'' \emph{IEEE transactions on neural networks and learning
  systems}, vol.~30, no.~9, pp. 2805--2824, 2019.

\bibitem{tsipras2019robustness}
D.~Tsipras, S.~Santurkar, L.~Engstrom, A.~Turner, and A.~Madry, ``Robustness
  may be at odds with accuracy,'' in \emph{International Conference on Learning
  Representations}, no. 2019, 2019.

\bibitem{woods2019reliable}
W.~Woods, J.~Chen, and C.~Teuscher, ``Reliable classification explanations via
  adversarial attacks on robust networks,'' \emph{arXiv preprint
  arXiv:1906.02896}, 2019.

\bibitem{fong2019understanding}
R.~Fong, M.~Patrick, and A.~Vedaldi, ``Understanding deep networks via extremal
  perturbations and smooth masks,'' in \emph{Proceedings of the IEEE
  International Conference on Computer Vision}, 2019, pp. 2950--2958.

\bibitem{fong2017interpretable}
R.~C. Fong and A.~Vedaldi, ``Interpretable explanations of black boxes by
  meaningful perturbation,'' in \emph{Proceedings of the IEEE International
  Conference on Computer Vision}, 2017, pp. 3429--3437.

\bibitem{etmann2019connection}
C.~Etmann, S.~Lunz, P.~Maass, and C.-B. Sch{\"o}nlieb, ``On the connection
  between adversarial robustness and saliency map interpretability,''
  \emph{arXiv preprint arXiv:1905.04172}, 2019.

\bibitem{xu2018structured}
K.~Xu, S.~Liu, P.~Zhao, P.-Y. Chen, H.~Zhang, Q.~Fan, D.~Erdogmus, Y.~Wang, and
  X.~Lin, ``Structured adversarial attack: Towards general implementation and
  better interpretability,'' \emph{arXiv preprint arXiv:1808.01664}, 2018.

\bibitem{papernot2016limitations}
N.~Papernot, P.~McDaniel, S.~Jha, M.~Fredrikson, Z.~B. Celik, and A.~Swami,
  ``The limitations of deep learning in adversarial settings,'' in \emph{2016
  IEEE European symposium on security and privacy (EuroS\&P)}.\hskip 1em plus
  0.5em minus 0.4em\relax IEEE, 2016, pp. 372--387.

\bibitem{elliott2019adversarial}
A.~Elliott, S.~Law, and C.~Russell, ``Adversarial perturbations on the
  perceptual ball,'' \emph{arXiv preprint arXiv:1912.09405}, 2019.

\bibitem{ghorbani2019towards}
A.~Ghorbani, J.~Wexler, J.~Y. Zou, and B.~Kim, ``Towards automatic
  concept-based explanations,'' in \emph{Advances in Neural Information
  Processing Systems}, 2019, pp. 9277--9286.

\bibitem{zhou2018interpretable}
B.~Zhou, Y.~Sun, D.~Bau, and A.~Torralba, ``Interpretable basis decomposition
  for visual explanation,'' in \emph{Proceedings of the European Conference on
  Computer Vision (ECCV)}, 2018, pp. 119--134.

\bibitem{yosinski2015understanding}
J.~Yosinski, J.~Clune, A.~Nguyen, T.~Fuchs, and H.~Lipson, ``Understanding
  neural networks through deep visualization,'' \emph{arXiv preprint
  arXiv:1506.06579}, 2015.

\bibitem{nguyen2017plug}
A.~Nguyen, J.~Clune, Y.~Bengio, A.~Dosovitskiy, and J.~Yosinski, ``Plug \& play
  generative networks: Conditional iterative generation of images in latent
  space,'' in \emph{Proceedings of the IEEE Conference on Computer Vision and
  Pattern Recognition}, 2017, pp. 4467--4477.

\bibitem{bau2018gan}
D.~Bau, J.-Y. Zhu, H.~Strobelt, B.~Zhou, J.~B. Tenenbaum, W.~T. Freeman, and
  A.~Torralba, ``Gan dissection: Visualizing and understanding generative
  adversarial networks,'' \emph{arXiv preprint arXiv:1811.10597}, 2018.

\bibitem{kingma2014adam}
D.~P. Kingma and J.~Ba, ``Adam: A method for stochastic optimization,''
  \emph{arXiv preprint arXiv:1412.6980}, 2014.

\bibitem{otsu1979threshold}
N.~Otsu, ``A threshold selection method from gray-level histograms,''
  \emph{IEEE transactions on systems, man, and cybernetics}, vol.~9, no.~1, pp.
  62--66, 1979.

\bibitem{keys1981cubic}
R.~Keys, ``Cubic convolution interpolation for digital image processing,''
  \emph{IEEE transactions on acoustics, speech, and signal processing},
  vol.~29, no.~6, pp. 1153--1160, 1981.

\bibitem{poursaeed2018generative}
O.~Poursaeed, I.~Katsman, B.~Gao, and S.~Belongie, ``Generative adversarial
  perturbations,'' in \emph{Proceedings of the IEEE Conference on Computer
  Vision and Pattern Recognition}, 2018, pp. 4422--4431.

\bibitem{zhang2019interpreting}
T.~Zhang and Z.~Zhu, ``Interpreting adversarially trained convolutional neural
  networks,'' \emph{ICML}, 2019.

\bibitem{bertalmio2000image}
M.~Bertalmio, G.~Sapiro, V.~Caselles, and C.~Ballester, ``Image inpainting,''
  in \emph{Proceedings of the 27th annual conference on Computer graphics and
  interactive techniques}.\hskip 1em plus 0.5em minus 0.4em\relax ACM
  Press/Addison-Wesley Publishing Co., 2000, pp. 417--424.

\bibitem{chen2018sketchygan}
W.~Chen and J.~Hays, ``Sketchygan: Towards diverse and realistic sketch to
  image synthesis,'' in \emph{Proceedings of the IEEE Conference on Computer
  Vision and Pattern Recognition}, 2018, pp. 9416--9425.

\end{thebibliography}
%




%




\begin{IEEEbiography}[{\includegraphics[width=1in,height=1.25in,clip,keepaspectratio]{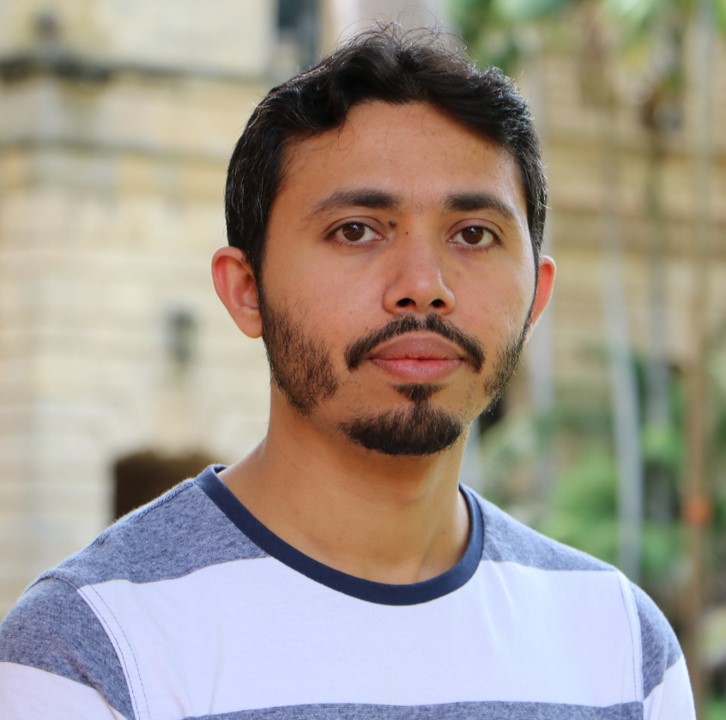}}]{Naveed Akhtar} is an Assistant Professor of Computer Science at The University of Western Australia (UWA). He received his PhD in Computer Vision from UWA and Master degree in Computer Science from Hochschule Bonn-Rhein-Sieg, Germany. He has served as a Research Fellow at UWA and the Australian National University in the past. His research is published regularly by the prestigious sources of his field. His research interests include adversarial deep learning, multiple object tracking, action recognition,  and hyperspectral image analysis.
\end{IEEEbiography}

\begin{IEEEbiography}[{\includegraphics[width=1in,height=1.25in,clip,keepaspectratio]{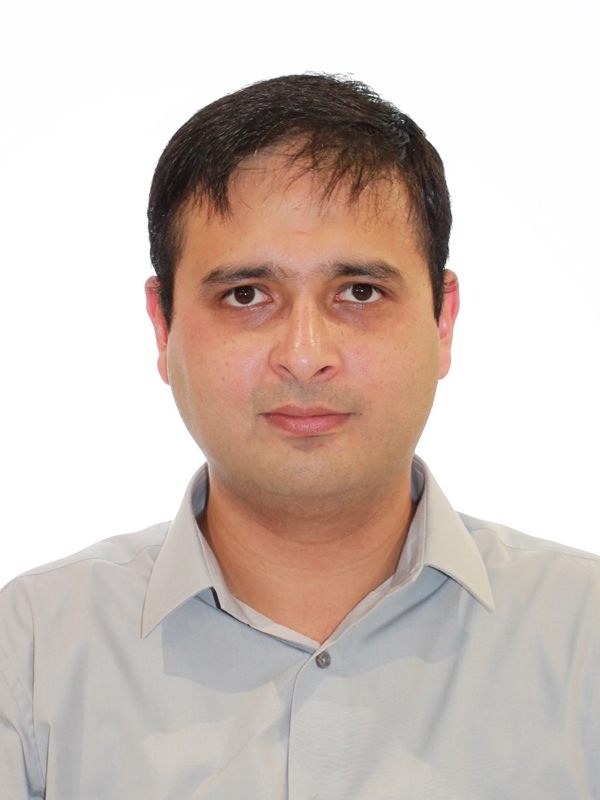}}]{Mohammad A. A. K. Jalwana} is currently working towards the PhD degree  at  The  University  of  Western  Australia (UWA) under the supervision of Prof. Ajmal Mian, Prof. Mohammed Bennamoun and  Dr.  Naveed  Akhtar.  He  is  a  recipient  of scholarship for international research fees (SIRF). His research interests are adversarial deep  learning  and  its  application in  computer vision, 3D localization, image captioning and medical image processing.
\end{IEEEbiography}

\begin{IEEEbiography}[{\includegraphics[width=1in,height=1.25in,clip]{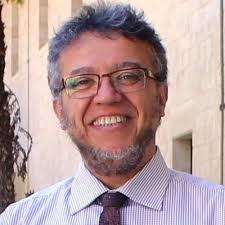}}]{MOHAMMED BENNAMOUN} is Winthrop Professor in the Department of Computer Science and Software Engineering at UWA and is a researcher in computer vision, machine/deep learning, robotics, and signal/speech processing. He has published 4 books (available on Amazon, 1 edited book, 1 Encyclopedia article, 14 book chapters, 120+ journal papers, 250+ conference publications, 16 invited \& keynote publications. His h-index is 54 and his number of citations is 12,500+ (Google Scholar). He was awarded 65+ competitive research grants, from the Australian Research Council, and numerous other Government, UWA and industry Research Grants. He successfully supervised +26 PhD students to completion. He won the Best Supervisor of the Year Award at QUT (1998), and received award for research supervision at UWA (2008 \& 2016) and Vice-Chancellor Award for mentorship (2016).  He delivered conference tutorials at major conferences, including: IEEE Computer Vision and Pattern Recognition (CVPR 2016), Interspeech 2014, IEEE International Conference on Acoustics Speech and Signal Processing (ICASSP) and European Conference on Computer Vision (ECCV). He was also invited to give a Tutorial at an International Summer School on Deep Learning (DeepLearn 2017).
\end{IEEEbiography}

\begin{IEEEbiography}[{\includegraphics[width=1in,height=1.25in,clip,keepaspectratio]{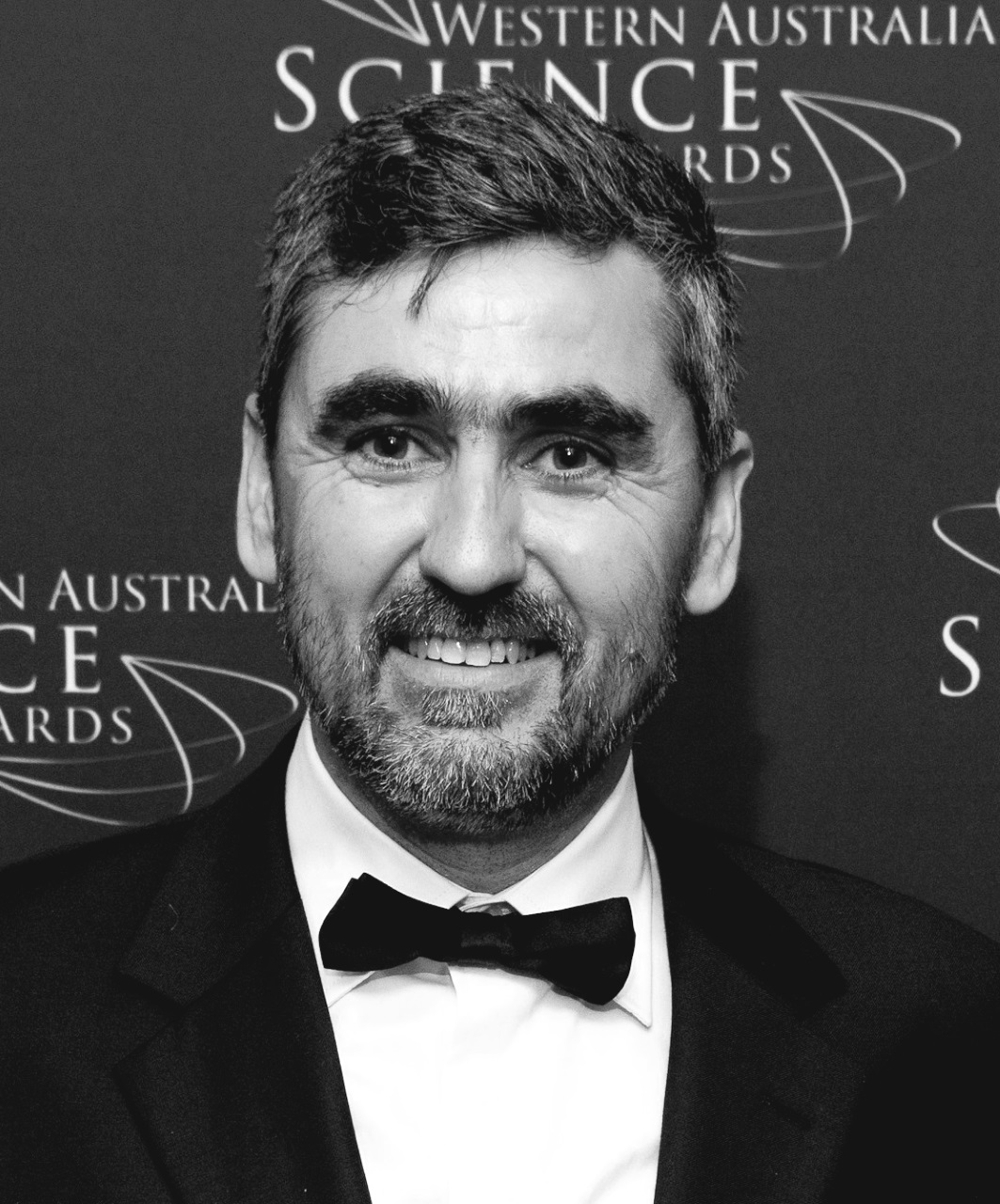}}]{Ajmal Mian} is a Professor of Computer Science at The University of Western Australia. He is the recipient of two prestigious national fellowships from the Australian Research Council and several awards including the West Australian Early  Career  Scientist  of  the  Year 2012, Excellence in Research Supervision, EH Thompson Award, ASPIRE Professional Development Award,  Vice-chancellors  Mid-career  Award,  Outstanding Young Investigator Award, the Australasian Distinguished Dissertation Award  and  various  best  paper  awards. He is an Associate Editor of IEEE Transactions on Neural Networks \& Learning Systems, IEEE Transactions on Image Processing and the Pattern Recognition journal. He served as the General Chair for DICTA 2019 and ACCV 2018. His  research  interests  are  in computer  vision,  deep  learning,  shape  analysis,  face  recognition, human action recognition and video analysis.
\end{IEEEbiography}




\end{document}